\documentclass[letterpaper,10pt,journal]{IEEEtran}

\usepackage{cite}
\usepackage{amsmath,amssymb,mathtools}
\usepackage{amsthm}
\usepackage{bm}
\usepackage{graphicx}
\usepackage{microtype}
\usepackage{booktabs}
\usepackage{array}
\usepackage{tabularx}
\usepackage{url}
\usepackage{xcolor}
\usepackage[skins,breakable]{tcolorbox}
\usepackage[hidelinks]{hyperref}

\definecolor{thmback}{RGB}{241,244,248}
\definecolor{thmrule}{RGB}{72,98,138}
\newtheoremstyle{paperthm}{0pt}{0pt}{\normalfont}{0pt}{\bfseries}{.}{0.5em}{\thmname{#1}\thmnumber{ #2}\thmnote{ \normalfont(#3)}}
\theoremstyle{paperthm}
\newtheorem{definition}{Definition}
\newtheorem{assumption}{Assumption}
\newtheorem{proposition}{Proposition}
\newtheorem{corollary}{Corollary}
\newtheorem{remark}{Remark}
\tcbset{paperthmbox/.style={enhanced, breakable, colback=thmback,
  colframe=thmrule, boxrule=0pt, leftrule=1.4pt, arc=0pt, outer arc=0pt,
  boxsep=0pt, left=5pt, right=5pt, top=3.5pt, bottom=3.5pt, before skip=7pt plus
  2pt, after skip=7pt plus 2pt}}
\tcolorboxenvironment{definition}{paperthmbox}
\tcolorboxenvironment{assumption}{paperthmbox}
\tcolorboxenvironment{proposition}{paperthmbox}
\tcolorboxenvironment{corollary}{paperthmbox}
\tcolorboxenvironment{remark}{paperthmbox}

\title{Actuation and Control of Passive Tendon-Driven\\ Soft Extensions Through
Rigid Manipulator Motion}
\author{Ignacio~Cuiral-Zueco, Gonzalo~L\'opez-Nicol\'as, and Youcef~Mezouar\thanks{This work has
been submitted to the IEEE for possible publication.
Copyright may be transferred without notice, after which this version may no
longer be accessible.}\thanks{This work was supported in part by the French National Research Agency
(ANR) under the France 2030 program, reference ANR-25-PERO-0003 (PEPR
Robotique, PC DRMI). I.~Cuiral-Zueco and G.~L\'opez-Nicol\'as are with the Instituto de
Investigaci\'on en Ingenier\'ia de Arag\'on (I3A), University of Zaragoza,
Spain. Y.~Mezouar is with Universit\'e Clermont Auvergne, Clermont Auvergne INP,
CNRS, Institut Pascal, F-63000 Clermont-Ferrand, France. Corresponding author:
I.~Cuiral-Zueco (e-mail: ignaciocuiral@unizar.es).}}
\begin{document}
\maketitle

\begin{abstract}
Industrial manipulators offer reach and precision but can damage nearby
structures in confined spaces. Soft robots provide compliant contact but
typically have limited reach. Rigid--soft systems combine their advantages,
although they often require dedicated actuation and control hardware. We propose
actuating a passive tendon-driven soft extension through manipulator joint
motion. As the robot moves, the exterior spans of tendons anchored to upstream
links change, altering tendon tension and deforming the extension.
Inspection through narrow apertures, for example, constrains the motion of
the end effector (EE). We develop a controller that regulates the soft tip
while preserving such EE constraints.
A differential model separates rigid transport from soft deformation,
combining analytic arm and tendon kinematics with a locally identified
span-to-deformation map. We also derive a local upper bound on the number
of independent deformation directions transmitted by the tendons under
the EE constraint, and conditions for the exponential convergence of the
tip error. A quadratic program with control barrier functions imposes
obstacle clearance, fulcrum limits, and model-validity conditions.
Simulations and real experiments show tip and shape regulation under free,
position-constrained, and fulcrum-constrained EE motion.
\end{abstract}

\begin{IEEEkeywords}
Continuum robots, hybrid rigid--soft robots, tendon-driven mechanisms, redundant
manipulators, task-priority control, inspection robotics.
\end{IEEEkeywords}

\section{Introduction}
\label{sec:introduction}

\begin{figure}[t]
    \centering
    \includegraphics[width=\columnwidth]{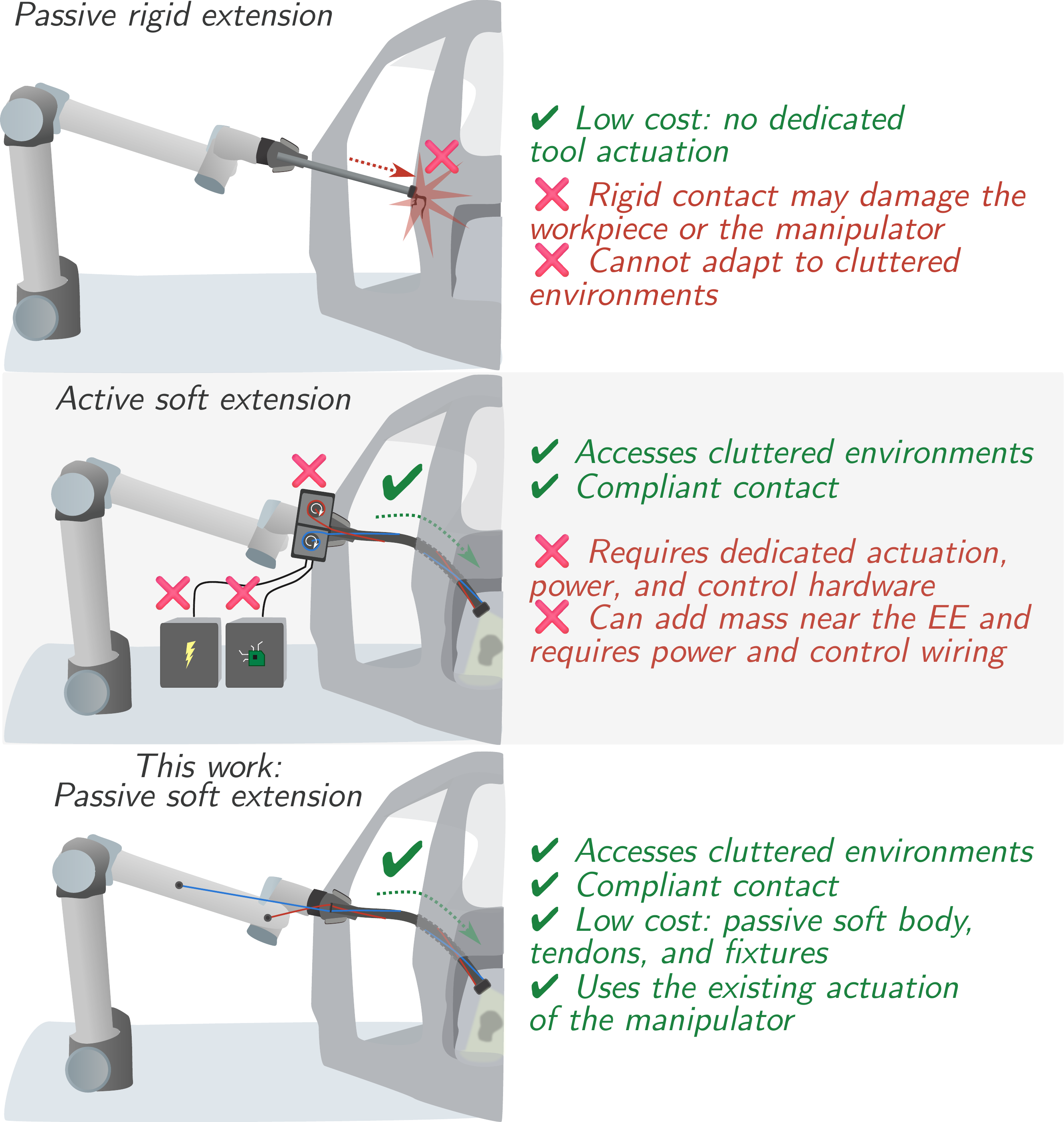}
    \vspace{0.55em}
    \footnotesize
    \setlength{\tabcolsep}{2pt}
    \renewcommand{\arraystretch}{1.05}
    \caption{Three robot
    \textit{extension} approaches for confined access: a passive rigid
    extension, an active soft extension, and the passive soft extension proposed
    here. The active design requires dedicated actuation, power, and control
    hardware, whereas the proposed design uses manipulator joint motion to
    actuate its tendons.}
    \label{fig:motivation}
\end{figure}

\IEEEPARstart{I}{ndustrial} manipulators are widely available, offer large
workspaces and precise positioning, and come with mature control
infrastructure. Their rigidity, however,
can pose safety risks and cause damage when operating in confined spaces. On the
other hand, soft and continuum robots can deform around obstacles and provide
compliant contact with their surroundings
\cite{webster2010,burgner2015,aner2026}, but typically have limited reach.
Mounting a soft robot at the end effector (EE) of a rigid manipulator combines
the rigid arm's reach with the soft robot's compliant access to confined
spaces. Existing rigid--soft combinations, however, typically require dedicated
actuation and control hardware for the soft robot, increasing integration
effort and adding mass near the EE \cite{camarillo2008,renda2018} (see
Fig.~\ref{fig:motivation}).

\begin{figure*}[!t]
    \centering
    \includegraphics[width=\textwidth]{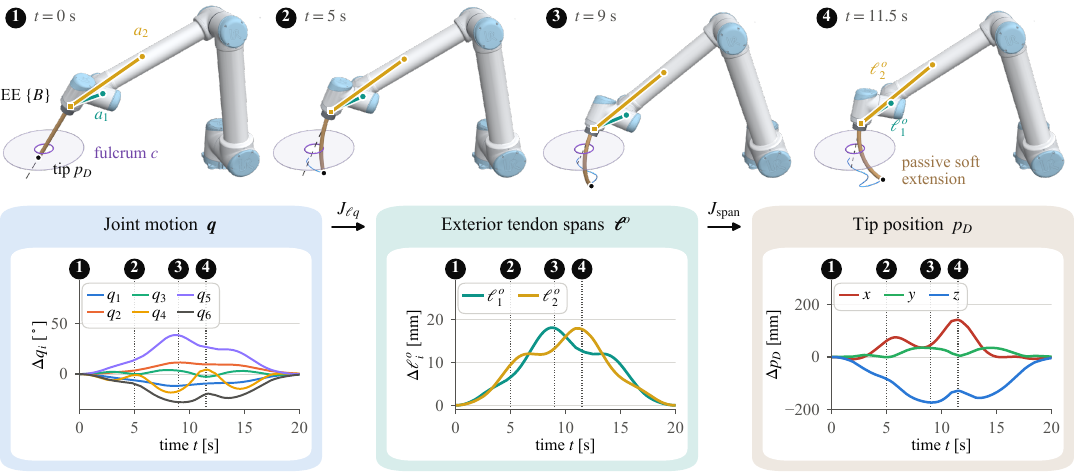}
    \caption{Manipulator-driven actuation of a passive soft extension in a UR10
    simulation. Top: four states under a laparoscopic fulcrum constraint. Two
    tendons run from anchors on the forearm and wrist (circles) to entry points
    at the base of the soft extension. The tip is black, and its path is blue.
    Bottom: corresponding joint angles, exterior tendon spans, and tip position.
    The spans change by up to $18\,\mathrm{mm}$ and the tip moves by up to
    $170\,\mathrm{mm}$.}
    \label{fig:mechanism}
\end{figure*}

We instead propose using the joint motion of the rigid manipulator (hereafter,
the robot) to actuate a passive tendon-driven soft extension (hereafter, the
extension). The extension is mounted at the robot's EE, and each tendon runs
from an anchor on an upstream arm link to an entry point at the extension's base
(Fig.~\ref{fig:mechanism}). We call the portion of each tendon between its
anchor and entry point its \emph{exterior span}. As the robot moves, changes in
these spans alter tendon tension and deform the extension. The robot can thus
position and deform the extension using its existing joints, and the extension
needs no dedicated actuators, sensors, or electronics.

This approach is particularly relevant to the inspection of hard-to-reach
spaces, where EE motion is often constrained to ensure proper insertion and
safety. For example, an inspection aperture may require a fixed EE position
while allowing reorientation, whereas a laparoscopic port permits pivoting and
axial insertion but prevents lateral motion at the entry point. Both cases
impose \emph{EE constraints}, which can be expressed at the velocity level as
\begin{equation}
    J_c(\bm q)\dot{\bm q}=0,
    \label{eq:intro_constraint}
\end{equation}
where $\bm q$ denotes the joint angles and $J_c$ is the constraint Jacobian.
Joint velocities that satisfy~\eqref{eq:intro_constraint} lie in the null space
of $J_c$ \cite{nakamura1987task}. We develop a modeling and control framework
that uses this \emph{null-space motion} to actuate the tendons and regulate the
soft tip while preserving the task-relevant EE constraint. The deformation
directions available for control therefore depend on the EE constraints and the
tendon and robot configuration.

Our main contributions are:

\begin{enumerate}
    \item \textbf{Manipulator-driven actuation of a passive soft extension.}
    We propose using rigid manipulators’ joint motions to actuate passive soft
    extensions through tendons anchored to upstream manipulator links, without
    requiring dedicated actuation or control hardware for the extensions. This
    approach enables low-cost and efficient integration of soft robotic capabilities into
    existing industrial robots and control infrastructure.

    \item \textbf{A differential model separating rigid transport from soft
    deformation.} We define a differential model relating joint velocity to
    soft-tip velocity. The model separates the \emph{transport} of the extension
    from its coupled tendon-driven deformation, where transport denotes its
    translation and rotation with the EE at fixed shape. We compute transport
    and tendon transmission analytically and propose a data-driven approach to
    identify the local span-to-deformation map. The decomposition also provides
    a local upper bound on the rank of the constrained deformation Jacobian
    (Section~\ref{sec:analysis}), which limits the number of independent
    deformation directions transmitted by the tendons. This bound accounts for
    the number of taut tendons, anchor placement, soft-body mechanics, and the
    null-space motion allowed by the EE constraint.

    \item \textbf{Constrained tip control with safety and model-validity
    conditions.} We treat the EE constraint as the primary task and use a damped
    least-squares control law to regulate the soft tip in its null space. For
    the nominal control law on the constraint manifold, we prove that a weighted
    tip-error Lyapunov function is nonincreasing. The tip error converges
    exponentially when the constrained tip Jacobian satisfies the rank and
    spectral conditions of Corollary~\ref{cor:exponential}. We also formulate a
    quadratic program using control barrier functions (CBFs) to maintain
    clearance between obstacles and the extension, respect the
    geometric limits of the fulcrum, and keep the spans within the identified
    span domain.

    \item \textbf{Simulation and experimental validation.} We evaluate the
    approach in simulation and on a physical UR10 robot arm. The simulation uses a
    Cosserat-rod model of the extension, and its code and interactive
    browser-based implementation are publicly available.\footnote{\raggedright
    Code: \url{https://github.com/nachocz/passive_soft_extension_control}.\\ Interactive pages:
    \url{https://ignaciocuiral.com/passive_soft_extension_control/}.} Six physical experiments
    use two low-cost handmade soft extensions with parallel and helical tendon
    routing to evaluate soft-tip control with a free EE, a position-constrained
    EE, and a fulcrum at a finite aperture (\textbf{see the accompanying
    video}).
\end{enumerate}

\section{Related Work}
\label{sec:related}

This work combines the actuation architectures and task-priority control of
rigid--soft systems with the local deformation model approach from deformable-object shape
control. Table~\ref{tab:related} compares the actuation and feedback
characteristics of these approaches with those of the passive soft extension.

\begin{table}[t]
\centering
\caption{Actuation and feedback characteristics of related approaches and of the
passive soft extension.}
\label{tab:related}
\footnotesize
\setlength{\tabcolsep}{2.2pt}
\renewcommand{\arraystretch}{1.15}

\begin{tabularx}{\columnwidth}{ @{}
>{\hsize=1.65\hsize\raggedright\arraybackslash}X
>{\hsize=0.70\hsize\centering\arraybackslash}X
>{\hsize=0.70\hsize\centering\arraybackslash}X
>{\hsize=0.95\hsize\centering\arraybackslash}X
>{\hsize=1.00\hsize\raggedright\arraybackslash}X @{}}
\toprule
Approach & Actuator-free & Shape regulation & Industrial manipulator interface &
Actuated by \\
\midrule

Continuum / tendon-driven
\cite{webster2010,camarillo2008}
& no & yes & no & own actuators \\

Hybrid rigid--continuum
\cite{yu2022,wilkening2017development}
& no & yes & yes & own actuators \\

Passive adaptive tools
\cite{chan2026,chan2025passive}
& yes & no$^{\ast}$ & yes & contact / arm motion \\

Deformable-object shape control
\cite{navarro2013model,cuiral2023multiscale}
& --$^{\dagger}$ & yes & yes & grasp motion \\

\textbf{Passive soft extension (this work)}
& \textbf{yes} & \textbf{yes} & \textbf{yes} & \textbf{arm joint motion} \\

\bottomrule
\end{tabularx}

\vspace{3pt}
\parbox{\columnwidth}{\footnotesize Actuator-free: no dedicated actuators at or
near the tool. Shape regulation: continuous feedback regulation of selected
shape features, such as tip and body-point positions. Industrial manipulator
interface: operated through the control interface of an existing industrial
manipulator. $^{\ast}$Configuration changes through contact or discrete arm
motions. $^{\dagger}$The deformable body is a workpiece rather than a tool.}
\end{table}

\subsection{Rigid--Soft Systems and Redundancy Resolution}
\label{subsec:rw_rigid_soft}

Continuum and soft manipulators deform along their length, which makes them
suitable for confined access \cite{webster2010,burgner2015,aner2026} and for
navigation among multiple obstacles \cite{ma2023}. Many of them are
tendon-driven \cite{camarillo2008,rucker2011statics}, including modular designs
with tendon motors embedded in each segment \cite{castledine2019}. Control
methods range from mechanics-based models \cite{dellasantina2023model} to the
online estimation of the Jacobian \cite{yip2014model}. Continuum robots have
also been mounted on aerial platforms \cite{chien2023}. Hybrid systems mount a
continuum section on a rigid arm to combine the compliance of the continuum
section with the reach of the arm, for example for pipe inspection \cite{yu2022}
or for surgical access to bone lesions \cite{wilkening2017development}. In the
arm-mounted systems, the manipulator positions the continuum section, and a
separate actuation module with its own motors deforms it. Recent designs reduce
the number of tendon actuators \cite{nino2025,yi2025position} but still drive
each remaining tendon with a motor.

Passive end effectors have no actuators of their own and change configuration
through the motion of the manipulator or through contact with the environment
\cite{chan2026}. A passive adaptive gripper, for example, closes when the
manipulator pushes it onto an object and opens at a prescribed release
orientation \cite{chan2025passive}. Such devices typically switch between
configurations rather than regulating their shape continuously by feedback.

A redundant manipulator retains joint motion in the null space of its primary
task. Redundancy-resolution methods assign this motion to secondary objectives,
from the gradient-projection method \cite{liegeois1977automatic} to
task-priority formulations with several tasks
\cite{nakamura1987task,siciliano1991}. Task-priority methods also represent the
remote center of motion of laparoscopic robots as a task constraint
\cite{aghakhani2013task}, and hierarchical quadratic programming extends the
priority hierarchy to inequality constraints \cite{escande2014hierarchical}.
Control barrier functions impose safety constraints as inequalities on the
control input \cite{ames2017control} and have recently been applied to soft
robots \cite{wong2025}.

\subsection{Shape Control of Deformable Objects}
\label{subsec:rw_shape}

Several reviews cover the robotic manipulation of deformable objects, in
particular of deformable linear objects (DLOs) such as cables and wires
\cite{sanchez2018robotic,zhu2022challenges,caporali2026survey}, and a taxonomy
classifies the associated shape-control problems \cite{cuiral2024taxonomy}. In
shape control (also called shape servoing), one or more robots move the grasp
points of an object to regulate its shape features, often with visual feedback.
Since accurate deformation models are difficult to obtain, many methods estimate
a local deformation Jacobian between grasp motion and shape-feature motion
online from measurements
\cite{navarro2013model,navarro2016automatic,lagneau2020automatic}. Proposed
shape features include the Fourier coefficients of a contour
\cite{navarro2018fourier}, the contour itself \cite{cuiral2023contour}, and 3-D
shapes aligned at several scales by Procrustes analysis
\cite{cuiral2023multiscale}.

Data-driven approaches include global models of large DLO deformations
\cite{yu2023global} and the direct control of DLO shape trajectories
\cite{cuiral2026data}. Planning-based approaches combine such models with
constraints from the environment, either by interleaving prediction, planning,
and control \cite{mcconachie2020manipulating} or by planning sets of paths that
satisfy the task constraints \cite{huang2023dom}. Shape control has also been
extended to co-manipulation tasks that involve large deformations
\cite{almaghout2024comanipulation}.

As in Jacobian-based shape control, our controller estimates a local deformation
Jacobian through probing motions and updates it online using measured deformations.
However, unlike in conventional shape control, where an object is deformed
through motion of its grasped points, our soft extension actuation is produced
by changes in the exterior tendon spans that the EE constraint allows.

\section{Modeling}
\label{sec:model}

\subsection{Notation}

Bold lower-case symbols denote coordinate vectors and task errors, upper-case
symbols denote matrices, and $I$ is the identity matrix. For a vector $x$,
$\|x\|$ is its Euclidean norm, and $[x]_\times$ is the skew-symmetric matrix
such that $[x]_\times y=x\times y$. For a matrix $J$, $J^+$ is its
Moore--Penrose pseudoinverse, and $\operatorname{rank}$ and
$\operatorname{range}$ denote the rank and the column space of a matrix. The
smallest eigenvalue of a symmetric matrix is $\lambda_{\min}(\cdot)$, and
$M\succ0$ means that $M$ is positive definite. Rigid-body poses are elements of
$SE(3)$, and a twist $\mathcal V=(v,\omega)\in\mathbb R^6$ stacks the linear and
angular velocities of a rigid body. The Appendix lists all symbols.

\subsection{General Model Description}

\begin{figure}[ht]
    \centering
    \includegraphics[width=\columnwidth]{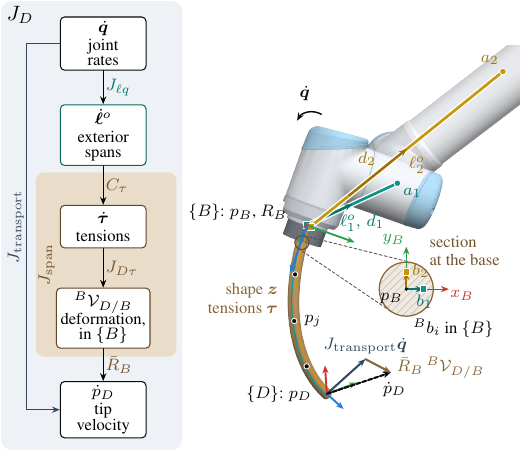}
    \caption{Notation. Left: the Jacobian $J_D$ (gray box) maps joint rates to
    the motion of the distal frame through two parallel paths whose
    contributions add, as in \eqref{eq:JD}. One path is the transport
    $J_{\mathrm{transport}}$ of the soft extension with the EE at fixed shape.
    The other, $\bar R_BJ_{\mathrm{span}}J_{\ell q}$, runs through the spans,
    tensions, and relative twist. Only its factor $J_{\mathrm{span}}$ (shaded)
    depends on the mechanics of the soft extension. Selecting the translational
    rows of $J_D$ gives the tip-position Jacobian $J_t=SJ_D$. Right: the
    simulated UR10 at $t=10\,\mathrm{s}$ in Fig.~\ref{fig:mechanism}. Joint
    motion $\dot{\bm q}$ moves the anchors $a_i$ relative to the entry points
    $b_i$ on the EE frame $\{B\}$ and changes the exterior spans $\ell^o_i$
    along the unit vectors $d_i$. The inset magnifies the cross-section of the
    soft extension at its base (ring), with the entry points at ${}^{B}b_i$ in
    $\{B\}$. The tendons continue inside the soft extension to its tip (drawn
    apart for visibility), and their tensions $\bm\tau$ deform the soft
    extension, whose shape $\bm z$ determines the distal frame $\{D\}$ and the
    body points $p_j$. The tip velocity is the translational part of the sum of
    transport, $J_{\mathrm{transport}}\dot{\bm q}$, and deformation, $\bar
    R_B\,{}^{B}\mathcal V_{D/B}$. Frames: $x$ red, $y$ green, $z$ blue.}
    \label{fig:notation}
\end{figure}

\begin{figure*}[t]
    \centering
    \includegraphics[width=\textwidth]{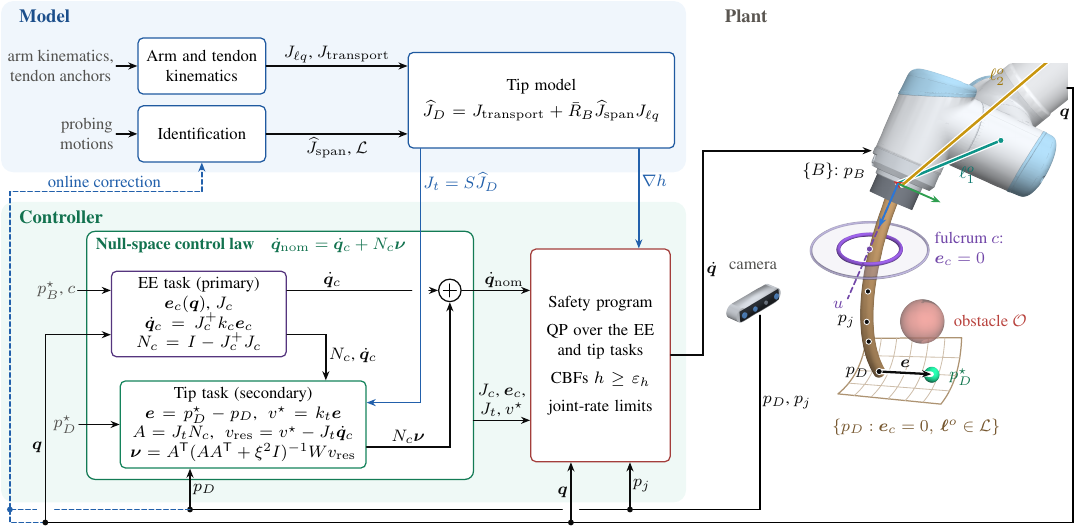}
    \caption{Overview of the method. Model (top): the arm kinematics and the
    measured tendon anchors give the analytic terms $J_{\ell q}$ \eqref{eq:Jlq}
    and $J_{\mathrm{transport}}$ \eqref{eq:Jtransport}, and the
    span-to-deformation map $\widehat J_{\mathrm{span}}$ and its span domain $\mathcal L$ are
    identified
    from probing motions and corrected online. Together they form the identified
    model $\widehat J_D$ \eqref{eq:JD_identified}. Controller (bottom): in the
    null-space control law \eqref{eq:hierarchical_law}, the EE task (primary)
    computes the restoring velocity $\dot{\bm q}_c$ and the projector $N_c$ from
    the error $\bm e_c$ about its reference, $p_B^\star$ or $c$
    (Table~\ref{tab:ee_constraints}). The tip task (secondary) uses both to
    compute the null-space motion $N_c\bm\nu$ \eqref{eq:damped_law}. When
    barriers are defined, the weighted quadratic program \eqref{eq:safety_qp}
    can replace $\dot{\bm q}_{\mathrm{nom}}$ to satisfy the barrier conditions.
    Plant (right): the robot provides $\bm q$, and a camera measures the tip
    $p_D$ and the body points $p_j$. In this example, the EE constraint is a
    fulcrum (purple) through the point $c$ of an aperture, and the grid
    schematically indicates the tip positions reachable with $\bm e_c=0$ and
    spans inside $\mathcal L$.}
    \label{fig:overview}
\end{figure*}

Our proposed model involves an analytic part, computed from the manipulator
geometry and the tendon anchor positions, and a data-driven part, which
maps tendon span changes to the deformation of the soft extension relative to the EE
(Fig.~\ref{fig:notation}). While mechanical models of a tendon-actuated soft body exist
\cite{tummers2023cosserat}, they require constitutive parameters and
boundary conditions that are difficult to identify in physical systems with
nonuniform mass, asymmetric tendon routing, heterogeneous material properties, and spatially
varying friction. We therefore identify this local map from probing data, as in
deformable-object shape control
\cite{cuiral2024taxonomy,navarro2013model,cuiral2025time}.

Consider an $n$-joint manipulator with joint angles $\bm q\in\mathbb R^n$ and EE
frame $\{B\}$, whose pose is $T_B(\bm q)=(R_B(\bm q),p_B(\bm q))\in SE(3)$, with
rotation $R_B\in SO(3)$ and position $p_B\in\mathbb R^3$. A passive soft
extension is rigidly attached to $\{B\}$ and ends at the distal frame $\{D\}$,
whose origin $p_D$ is the tip. It carries $m$ tendons, with their distal ends
fixed in the soft extension and their proximal ends anchored to links of the
manipulator upstream of the EE.

\subsection{Span Kinematics}
\label{subsec:rigid}

For tendon $i$, let $a_i(\bm q)$ be the world position of its anchor and
$b_i(\bm q)$ that of its entry point, where the tendon enters the soft
extension. The entry point is fixed in $\{B\}$ at ${}^{B}b_i$, and the exterior
span $\ell^o_i$ (span, for short) is the straight length of the tendon between
the two points,
\begin{equation}
    b_i(\bm q)=p_B(\bm q)+R_B(\bm q)\,{}^{B}b_i,
    \qquad
    \ell_i^o(\bm q)=\left\|a_i(\bm q)-b_i(\bm q)\right\|.
    \label{eq:span}
\end{equation}
The span vector $\bm\ell^o\in\mathbb R^m$ stacks the spans, and its rate is
\begin{equation}
    \dot{\bm\ell}^o=J_{\ell q}(\bm q)\dot{\bm q},
    \qquad
    J_{\ell q}:=\frac{\partial\bm\ell^o}{\partial\bm q}.
    \label{eq:Jlq_def}
\end{equation}
Let $d_i=(a_i-b_i)/\|a_i-b_i\|$ be the unit vector from the entry point $b_i$ to
the anchor $a_i$, $J_{a_i}^v$ the translational Jacobian of the anchor, and
$J_B$ the geometric Jacobian of the EE frame $\{B\}$. Its first three rows
$J_B^v$ give the linear velocity of $p_B$, and its last three rows $J_B^\omega$
the angular velocity of $\{B\}$, both expressed in the world frame.
Differentiating \eqref{eq:span} gives row $i$ of $J_{\ell q}$,
\begin{equation}
    [J_{\ell q}]_i
    =
    d_i^{\mathsf T}
    \left(
        J_{a_i}^{v}
        -J_B^v
        +[R_B\,{}^{B}b_i]_\times J_B^\omega
    \right).
    \label{eq:Jlq}
\end{equation}
Computing this matrix requires only the arm kinematics and the positions of the
anchors and entry points, and no model of the soft extension.

\subsection{Span-to-Deformation Map}
\label{subsec:soft}

A taut tendon runs from its anchor along its exterior span and then along an
interior path inside the soft extension to its distal end. If the tendon were
inextensible, its total length would remain constant, and a longer exterior span
would require a shorter interior path, that is, a deformation of the soft
extension. With finite axial stiffness, part of the span change stretches the
tendon instead.

To model both effects, let $\bm\tau\in\mathbb R^m$ be the tendon tensions and
${}^{B}\mathcal V_{D/B}\in\mathbb R^6$ the twist of $\{D\}$ relative to $\{B\}$,
expressed in $\{B\}$. The shape coordinates $\bm z$ are generalized coordinates
that describe the deformed configuration of the soft extension, such as the
curvatures and torsions of the segments of a discretized rod. A tendon is slack
when it transmits no tension and saturated when its tension has reached a
modeled upper limit, so that a further span change no longer changes it. For a
fixed active set of taut, unsaturated tendons, let
$K_a=\operatorname{diag}(k_{a,1},\ldots,k_{a,m})$, where $k_{a,i}$ is the axial
stiffness of tendon $i$ if it is active and zero otherwise, and let $J_{\ell z}$
map shape changes to changes of the interior tendon lengths. Since the tension
depends on the total stretch of the tendon, the linearized tension--stretch
relation of the tendons includes the length changes of both the exterior span
and the interior path,
\begin{equation}
    \dot{\bm\tau}
    =
    K_a
    \left(
        \dot{\bm\ell}^o
        +
        J_{\ell z}\dot{\bm z}
    \right).
    \label{eq:tension_linearization}
\end{equation}
Let $J_{z\tau}=\partial\bm z/\partial\bm\tau$ denote the local shape response to
the tendon tensions, as given by a statics model of tendon-driven continuum
robots such as that of \cite{rucker2011statics}. Substituting $\dot{\bm
z}=J_{z\tau}\dot{\bm\tau}$ into \eqref{eq:tension_linearization} and solving for
$\dot{\bm\tau}$ gives the span-to-tension map
\begin{equation}
    C_\tau
    :=
    \frac{\partial\bm\tau}{\partial\bm\ell^o}
    =
    \left(
        I-K_aJ_{\ell z}J_{z\tau}
    \right)^{-1}K_a .
    \label{eq:Ctau}
\end{equation}
When $K_a$ is invertible, \eqref{eq:Ctau} equals $(K_a^{-1}+G)^{-1}$, where
$G:=-J_{\ell z}J_{z\tau}$ is the compliance of the soft extension seen at the
tendons. The compliances of the tendons and of the soft extension thus add as
for springs in series. Inextensible tendons are the limit $K_a^{-1}\to0$, in
which $C_\tau$ tends to $G^{-1}$ and \eqref{eq:tension_linearization} reduces to
$\dot{\bm\ell}^o=-J_{\ell z}\dot{\bm z}$. With $J_{D\tau}$ mapping tension
changes to the relative distal twist, span changes produce the twist
\begin{equation}
    {}^{B}\mathcal V_{D/B}=J_{\mathrm{span}}\dot{\bm\ell}^o,
    \qquad
    J_{\mathrm{span}}:=J_{D\tau}C_\tau .
    \label{eq:Jspan}
\end{equation}
The controller depends on the mechanics of the soft extension only through
$J_{\mathrm{span}}$,
which Section~\ref{subsec:identified} identifies directly from data, whereas the
factorization \eqref{eq:Jspan} serves only the rank analysis of
Section~\ref{sec:analysis}.

\subsection{Tip Velocity}
\label{subsec:tip_velocity}

The twist of the distal frame is the sum of the transport imposed by the EE at
fixed shape and the deformation that the tendons produce relative to the EE. Let
\begin{equation}
    J_{\mathrm{transport}}
    =
    \begin{bmatrix}
        J_B^v-[p_D-p_B]_\times J_B^\omega\\
        J_B^\omega
    \end{bmatrix}
    \label{eq:Jtransport}
\end{equation}
be the Jacobian of $\{D\}$ as if $\{D\}$ were rigidly attached to $\{B\}$ in its
current position. It depends on $\bm q$ and, through $p_D-p_B$, on the current
shape, and is thus recomputed at every control step. With $\bar
R_B=\operatorname{diag}(R_B,R_B)$, which rotates a twist from $\{B\}$ to the
world frame, the distal twist is $J_D\dot{\bm q}$, where
\begin{equation}
    J_D
    =
    J_{\mathrm{transport}}
    +
        \bar R_BJ_{\mathrm{span}}J_{\ell q}.
    \label{eq:JD}
\end{equation}
Its translational rows, selected by $S$, form the tip-position Jacobian
$J_t=SJ_D$. The controller uses both terms of $J_D$, since the EE constraint
does not, in general, fix the orientation of the EE. Under the position
constraint, for example, joint motion can still rotate the EE, and this rotation
moves the tip even if the spans remain constant.

\subsection{Identified Model}
\label{subsec:identified}

During identification, the arm moves through probe configurations that vary the
exterior spans. A camera measures the tip and a few body points $p_j$ along the
soft extension. With the analytic terms $J_{\ell q}$ and
$J_{\mathrm{transport}}$ computed from the arm kinematics and the measured
anchor and entry points, the controller uses
\begin{equation}
    \widehat J_D
    =
    J_{\mathrm{transport}}
    +
    \bar R_B\widehat J_{\mathrm{span}}J_{\ell q},
    \label{eq:JD_identified}
\end{equation}
where $\widehat J_{\mathrm{span}}$ is fitted to the probing data and updated
online from the measured joint and tip motion. We apply Broyden's rank-one
secant update \cite{broyden1965class}. The fitted model also predicts the
positions ${}^{B}p_j(\bm\ell^o)$ of the body points, expressed in $\{B\}$, and
their sensitivities to the spans. Their world positions $p_j=p_B+R_B\,{}^{B}p_j$
then have the joint-space Jacobians
\begin{equation}
    J_{p_j}
    =
    J_B^v-[R_B\,{}^{B}p_j]_\times J_B^\omega
        +R_B\frac{\partial\,{}^{B}p_j}{\partial\bm\ell^o}J_{\ell q},
    \label{eq:shape_chain}
\end{equation}
which require no differentiation of a continuum-mechanics solver. Since the identification is local,
the model is restricted to the identified
span domain $\mathcal L=[\bm\ell^-,\bm\ell^+]$, which contains the exterior
spans covered by the probing motions. The span-domain barrier
$h_{\mathrm{span}}$ of Table~\ref{tab:barriers} constrains the spans to remain
inside this domain.

\section{Control}
\label{sec:control}

The nominal controller follows the task-priority formulation of
\cite{nakamura1987task}, in which the EE constraint is the primary task and the
tip is regulated in its null space using the model of Section~\ref{sec:model}
(Fig.~\ref{fig:overview}, bottom).

\subsection{EE Constraints}
\label{subsec:ee_constraints}

\begin{figure}[t]
    \centering
    \includegraphics[width=\columnwidth]{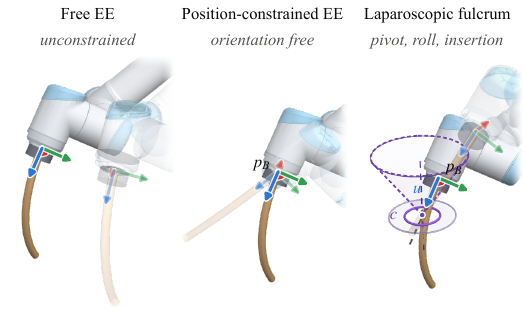}
    \caption{The three EE constraints on the simulated UR10. Each panel compares
    a reference EE pose (solid) with another pose that satisfies the constraint
    (faded, over a translucent arm). Free EE: the task constrains no joint
    motion. Position-constrained EE: $p_B$ is fixed and only
    the orientation can change. Laparoscopic fulcrum: the tool axis $u$ (the $z$
    axis of the EE) passes through the point $c$, the cone (dashed outline)
    shows the tilt limit of the cone barrier (Table~\ref{tab:barriers}), and
    the arm can pivot, roll, and insert.
    Table~\ref{tab:ee_constraints} gives the errors. Frames: $x$ red, $y$ green,
    $z$ blue.}
    \label{fig:ee_constraints}
\end{figure}

As representative cases, we consider a free EE, a position-constrained EE, and the ideal
laparoscopic
fulcrum (Fig.~\ref{fig:ee_constraints} and Table~\ref{tab:ee_constraints}). The fulcrum
requires the tool axis to pass through a fixed point $c$, the remote center of
motion in laparoscopic robotics \cite{aghakhani2013task}. Under this constraint, the arm can still
pivot
about $c$, rotate about the tool axis, and insert the tool.
This ideal constraint represents tasks in which a tool passes through a small
aperture with negligible lateral clearance, such as a trocar in surgery or an
access hole in an inspection task.

Let the constraint manifold be defined by $\bm e_c(\bm q)=0$, where $\bm e_c$
is the constraint error of Table~\ref{tab:ee_constraints}, and let the
constraint Jacobian be $J_c=-\partial\bm e_c/\partial\bm q$, so that
\begin{equation}
    \dot{\bm e}_c
    =
    -J_c(\bm q)\dot{\bm q}.
    \label{eq:constraint_error_rate}
\end{equation}
For the position constraint, for example, $J_c=J_B^v$. By
\eqref{eq:constraint_error_rate}, joint velocities that satisfy
\eqref{eq:intro_constraint} give $\dot{\bm e}_c=0$, so once the constraint is
satisfied, it is preserved.

As in the gradient-projection method of \cite{liegeois1977automatic}, the
orthogonal projector onto the null space of $J_c$ is
\begin{equation}
    N_c
    =
    I-J_c^+J_c.
    \label{eq:Nc}
\end{equation}
With $r_c=\operatorname{rank}J_c$, the range of $N_c$ has dimension $n-r_c$, and
every joint velocity tangent to the constraint manifold can be written as
$\dot{\bm q}=N_c\bm\nu$ with an arbitrary $\bm\nu\in\mathbb R^n$. To correct
small deviations from the constraint caused by initialization, kinematic model
error, and time discretization, the controller adds the restoring velocity
\begin{equation}
    \dot{\bm q}_c
    =
    J_c^+k_c\bm e_c,
    \label{eq:primary}
\end{equation}
with gain $k_c>0$. The nominal command is the sum of this correction and a
null-space motion that regulates the tip,
\begin{equation}
    \dot{\bm q}_{\mathrm{nom}}
    =
    \dot{\bm q}_c
    +
    N_c\bm\nu.
    \label{eq:hierarchical_law}
\end{equation}

\subsection{Tip Task in the Null Space}
\label{subsec:tip_task}

\begin{table}[t]
\centering
\caption{EE constraints (Fig.~\ref{fig:ee_constraints}).}
\label{tab:ee_constraints}
\footnotesize
\setlength{\tabcolsep}{5pt}
\renewcommand{\arraystretch}{1.2}
\begin{tabular}{@{}l>{$}l<{$}c@{}}
\toprule
Constraint & \multicolumn{1}{l}{Error $\bm e_c(\bm q)$} & $r_c$ \\
\midrule
Free EE & \multicolumn{1}{l}{none} & $0$ \\
Position-constrained EE & p_B^\star-p_B & $3$ \\
Laparoscopic fulcrum & (I-uu^{\mathsf T})(c-p_B) & $2$ \\
\bottomrule
\end{tabular}

\vspace{3pt}
\parbox{\columnwidth}{\footnotesize $p_B^\star$: held EE point; $c$: fulcrum
point; $u$: tool axis, the unit $z$ axis of $\{B\}$; $r_c$: rank of $J_c$. On
the experimental setup, the fulcrum is a finite aperture imposed by barriers only
($r_c=0$).}
\end{table}

The tip error and the proportional tip-velocity reference are
\begin{equation}
    \bm e
    =
    p_D^\star-p_D,
    \qquad
    v^\star
    =
    k_t\bm e,
    \label{eq:tip_error}
\end{equation}
where $p_D^\star$ is the target, $p_D$ the measured tip position, and $k_t>0$
the tip gain. We derive the control law for the tip position. For shape regulation,
$p_D$, $\bm e$, and $J_t$ stack the positions, errors, and Jacobians of the tip
and selected body points, and the derivation is unchanged. A null-space command
moves the tip by $J_tN_c\bm\nu$, and the
\emph{constrained tip Jacobian}
\begin{equation}
    A
    :=
    J_tN_c
    \label{eq:A}
\end{equation}
is thus the tip Jacobian restricted to the joint motions that the EE constraint
permits. Since the restoring velocity also moves the tip, by $J_t\dot{\bm q}_c$,
the null-space term tracks the residual reference $v_{\mathrm{res}}:=v^\star-J_t
\dot{\bm q}_c$. We compute the vector $\bm\nu$ in
\eqref{eq:hierarchical_law} with the damped least-squares inverse of
$A$ \cite{nakamura1986inverse,chiaverini1997singularity},
\begin{equation}
    \bm\nu
    =
    A^{\mathsf T}
    (AA^{\mathsf T}+\xi^2I)^{-1}
    Wv_{\mathrm{res}},
    \label{eq:damped_law}
\end{equation}
where $W\succ0$ weights the tip coordinates and $\xi>0$ is the damping, in
meters, since $A$ maps joint rates in rad/s to tip velocities in m/s. The
null-space tip velocity $A\bm\nu$ lies in $\operatorname{range}A$, and null-space
motion can therefore produce only the component of $v_{\mathrm{res}}$ in this
subspace.

\subsection{Safety and Model-Validity Barriers}
\label{subsec:safety}

\begin{figure}[t]
    \centering
    \includegraphics[width=\columnwidth]{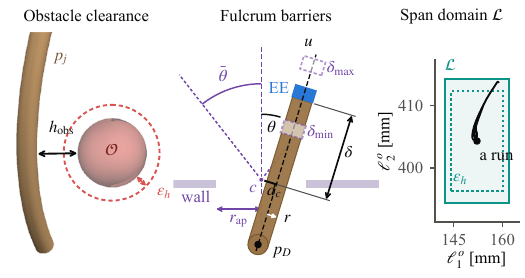}
    \caption{The barrier functions of Table~\ref{tab:barriers}; dashed lines
    mark the margins $h=\varepsilon_h$. Left: obstacle clearance between the
    surface of the soft extension (body points $p_j$, radius $r$) and the
    obstacle $\mathcal O$. Middle: fulcrum barriers in a section through the
    aperture (center $c$, radius $r_{\mathrm{ap}}$), for the aperture crossing
    of the tool axis $u$ ($h_{\mathrm{ap}}$), its tilt below the cone half-angle
    $\bar\theta$ ($h_{\mathrm{cone}}$), and the insertion depth $\delta$ between
    $\delta_{\min}$ and $\delta_{\max}$ ($h_{\mathrm{ins}}$; EE dashed at both
    limits). Right: the spans inside the identified span domain $\mathcal L$ in
    simulation, with the spans of one run.}
    \label{fig:barriers}
\end{figure}

The control law \eqref{eq:hierarchical_law} does not include inequality
conditions for other considerations like obstacles, insertion depth, or joint-rate limits. Except
for the joint-rate
limits, which are imposed as bounds, each condition is
written as a barrier function $h(\bm q)$ that must remain above a margin
$\varepsilon_h\ge0$. Table~\ref{tab:barriers} defines five such barriers,
which Fig.~\ref{fig:barriers} illustrates. Two of them concern obstacle
clearance and the span domain, and three describe the fulcrum. The fulcrum
barriers constrain where the tool axis crosses the aperture ($h_{\mathrm{ap}}$),
its tilt within a cone of half-angle $\bar\theta$ ($h_{\mathrm{cone}}$), and the
insertion depth ($h_{\mathrm{ins}}$). For the ideal fulcrum of
Table~\ref{tab:ee_constraints}, only the last two are needed, since its equality
constraint already requires the tool axis to pass through $c$. When any barrier is enabled, the
robot
command is the solution of the weighted quadratic program (QP)
\begin{equation}
\begin{aligned}
    \min_{\dot{\bm q}}\quad&
    \|J_t\dot{\bm q}-v^\star\|^2
    +
    w_c\|J_c\dot{\bm q}-k_c\bm e_c\|^2
    +
    \mu^2\|\dot{\bm q}\|^2
    \\
    \text{s.t.}\quad&
    \nabla h(\bm q)^{\mathsf T}\dot{\bm q}
    \ge
    -\alpha_h\bigl(h(\bm q)-\varepsilon_h\bigr)
    \quad\text{for all }h\in\mathcal H,
    \\
    &
        \dot{\bm q}^{-}\le\dot{\bm q}\le\dot{\bm q}^{+}.
\end{aligned}
\label{eq:safety_qp}
\end{equation}
Here, $\mathcal H$ is the set of barriers in Table~\ref{tab:barriers}, each
inequality is a CBF condition with rate $\alpha_h>0$ \cite{ames2017control},
$\mu>0$ weights the regularization of the joint rates, and $\dot{\bm q}^{\pm}$
are the joint-rate limits. In these conditions, the margin $\varepsilon_h$
creates a buffer for finite-step linearization and modeling errors. The objective
penalizes the residuals of the two tasks of
\eqref{eq:hierarchical_law}, and $w_c\gg1$ gives the EE residual a much larger
weight than the tip residual. As the QP uses finite task weights, it
only approximates the strict hierarchy of \eqref{eq:hierarchical_law}, even when
no inequality is active, whereas hierarchical QP solvers such as
\cite{escande2014hierarchical} impose the hierarchy exactly.

\begin{table}[t]
\centering
\caption{Barrier functions, $h\ge\varepsilon_h$ (Fig.~\ref{fig:barriers}).}
\label{tab:barriers}
\footnotesize
\setlength{\tabcolsep}{3pt}
\renewcommand{\arraystretch}{1.2}
\begin{tabularx}{\columnwidth}{@{}>{\raggedright\arraybackslash}X>{$}l<{$}>{$}l<{$}@{}}
\toprule
Barrier & \multicolumn{1}{l}{Function $h$} & \multicolumn{1}{l@{}}{Gradient $\nabla h^{\mathsf T}$}
\\
\midrule
Obstacle clearance &
h_{\mathrm{obs}}=\operatorname{dist}(\mathcal P,\mathcal O)-r & \eta^{\mathsf T}J_p \\
Fulcrum: aperture &
h_{\mathrm{ap}}=(r_{\mathrm{ap}}-r)\cos\theta-d_c & \text{from } J_B \\
Fulcrum: cone &
h_{\mathrm{cone}}=\bar\theta-\theta & \text{from } J_B \\
Fulcrum: insertion &
h_{\mathrm{ins}}=\min(\delta-\delta_{\min},\,\delta_{\max}-\delta) & \text{from } J_B \\
Span domain $\mathcal L$ &
h_{\mathrm{span}}=\min_i\{\ell^o_i-\ell^-_i,\,\ell^+_i-\ell^o_i\} & \pm[J_{\ell q}]_i \\
\bottomrule
\end{tabularx}

\vspace{3pt}
\parbox{\columnwidth}{\footnotesize $\mathcal P$: polyline through the body
points (Sections~\ref{subsec:sim_setup} and~\ref{subsec:exp_runs});
$\operatorname{dist}(\mathcal P,\mathcal O)$: distance from $\mathcal P$ to the
obstacle $\mathcal O$; $r$, $r_{\mathrm{ap}}$: radii
of the soft extension and of the aperture; $d_c$: distance from $c$ to the tool
axis $u$; $\theta$: tilt of $u$ from the axis of the aperture; $\bar\theta$:
half-angle of the cone; $\delta=u^{\mathsf T}(c-p_B)$: insertion depth;
$\mathcal L=[\bm\ell^-,\bm\ell^+]$: identified span domain; $\eta$: unit vector
from $\mathcal O$ to the point $p$ of $\mathcal P$ closest to it, whose
Jacobian $J_p$ is interpolated from \eqref{eq:shape_chain}.}
\end{table}

The gradient of the obstacle barrier follows from the body-point Jacobians
\eqref{eq:shape_chain}, whose span sensitivities come from the identified model.
By contrast, the fulcrum and span-domain gradients depend only on the arm
kinematics. Each minimum in Table~\ref{tab:barriers}
is split into separate inequalities, one per obstacle, per insertion limit, and
per lower or upper span bound, so that each inequality has a well-defined
gradient.

\section{Analysis}
\label{sec:analysis}

\subsection{Deformation Available Under the EE Constraint}
\label{subsec:authority}

\subsubsection{Tendon-Transmitted Deformation}
To characterize the deformation transmitted by the tendons separately from the
transport in the constrained tip Jacobian $A$, we restrict the
tendon-dependent part of the Jacobian to the null space of the EE constraint.

\begin{definition}[Constrained deformation Jacobian]
\label{def:deformation}
The constrained deformation Jacobian is
\begin{equation}
    J_{\mathrm{def}}
    :=
    J_{\mathrm{span}}J_{\ell q}N_c
    =
    J_{D\tau}C_\tau J_{\ell q}N_c.
    \label{eq:Jdef}
\end{equation}
It maps the null-space component of a joint velocity to the twist of $\{D\}$
relative to $\{B\}$ produced by the tendons.
\end{definition}

By \eqref{eq:Jspan}, \eqref{eq:JD}, and \eqref{eq:A}, the constrained tip
Jacobian separates into transport at fixed shape and tendon-transmitted
deformation,
\begin{equation}
    A
    =
    J_tN_c
    =
    SJ_{\mathrm{transport}}N_c
    +
    S\bar R_BJ_{\mathrm{def}}.
    \label{eq:A_decomposition}
\end{equation}

\begin{proposition}[Local rank bound for the constrained deformation Jacobian]
\label{prop:authority}
Let $m_a$ be the number of taut, unsaturated tendons in a fixed active set. Then
\begin{equation}
\begin{aligned}
    \operatorname{rank}J_{\mathrm{def}}
    &\le
    \min\left\{
        \operatorname{rank}J_{D\tau},
        m_a,
        \operatorname{rank}(J_{\ell q}N_c)
    \right\}
    \\
    &\le
    \min\left\{
        \operatorname{rank}J_{D\tau},
        m_a,
        \operatorname{rank}J_{\ell q},
        n-r_c
    \right\}.
\end{aligned}
\label{eq:rank_bound}
\end{equation}
\end{proposition}

\begin{IEEEproof}
The rank of a matrix product does not exceed the rank of any of its factors.
Applied to \eqref{eq:Jdef}, with
$\operatorname{rank}C_\tau\le\operatorname{rank}K_a=m_a$ by \eqref{eq:Ctau},
this gives the first inequality. The second follows from
$\operatorname{rank}(J_{\ell q}N_c)\le\min\{\operatorname{rank}J_{\ell
q},\operatorname{rank}N_c\}$ and $\operatorname{rank}N_c=n-r_c$.
\end{IEEEproof}

The bound \eqref{eq:rank_bound} combines mechanical and kinematic factors, and
its smallest factor limits the rank of $J_{\mathrm{def}}$. Its mechanical
factors are the rank of the
tension-to-twist map $J_{D\tau}$ and the number $m_a$ of taut, unsaturated
tendons. The
kinematic factors are the span
sensitivity $J_{\ell q}$, which depends on the anchor placement, and the
null-space dimension $n-r_c$, which counts the independent joint velocities that
preserve the EE constraint. An additional tendon increases $m_a$ only when it is
taut and unsaturated, and it increases $\operatorname{rank}(J_{\ell q}N_c)$ only
when null-space motion changes its span independently of the other spans.
Similarly, an additional null-space
direction increases this rank only when it produces an independent span-rate
direction.

\begin{corollary}[Transmission-null configuration]
\label{cor:transmission_null}
If $J_{\ell q}N_c=0$ at a configuration, then $J_{\mathrm{def}}=0$ there, so
null-space motion produces no tendon-transmitted deformation at that
configuration.
\end{corollary}

Corollary~\ref{cor:transmission_null} concerns only deformation transmitted
through the tendons. Configuration-dependent loads can deform the soft extension
even when $J_{\ell q}N_c=0$. Changing the orientation of its base, for example,
changes the gravitational load expressed in the body frame, and with it the
equilibrium shape, while the exterior spans remain constant.

\begin{remark}
\label{rem:local}
Proposition~\ref{prop:authority} is local. It bounds the rank of the
constrained deformation Jacobian at a configuration and does not establish
global controllability.
\end{remark}

\subsubsection{Design Implications}
Anchor placement determines $J_{\ell q}$ and thus the deformation directions
that the passive actuation can produce. For a prescribed EE
constraint, the
anchors should be placed so that null-space motion produces nonzero and
sufficiently independent changes of the exterior spans. In particular, an anchor
on the EE moves together with the entry point, which gives a zero row of
$J_{\ell q}$ and no contribution to tendon-driven deformation. Anchor
placements should therefore be evaluated from $J_{\ell q}N_c$ and, when the
mechanical maps are available, from $J_{\mathrm{def}}$.

\subsection{Convergence of the Nominal Controller}
\label{subsec:stability}

\begin{figure}[t]
    \centering
    \includegraphics[width=\columnwidth]{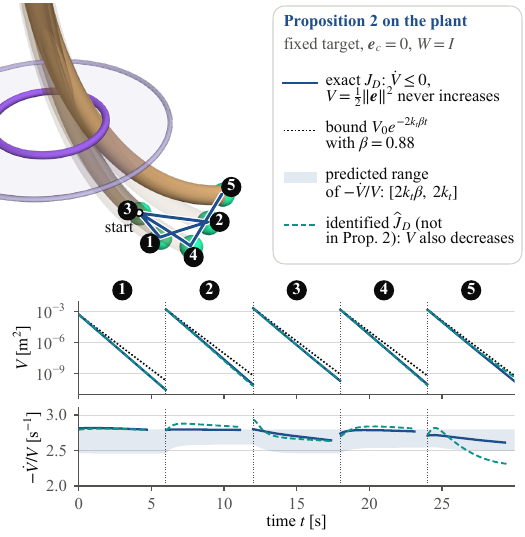}
    \caption{Convergence of the nominal controller
    \eqref{eq:on_manifold_controller} on the simulated plant for five
    targets, reached in sequence under the fulcrum, with the damped inverse
    ($k_t=1.4$, $\xi=0.04\,\mathrm{m}$), $W=I$, and no
    barriers. Along the run, $A$ keeps full row rank and $\beta$ stays between
    $0.88$ and $0.92$. Top:
    tip path through the numbered targets, over the soft extension at each
    target switch, and a legend of the plots. Middle: $V$ on a logarithmic
    vertical axis, with the exact Jacobian
    (solid) and with the identified model \eqref{eq:JD_identified} (dashed).
    The bound \eqref{eq:exp_convergence} (dotted) restarts at each target
    switch from the current value of $V$ and uses the smallest $\beta$ of the
    run. Bottom: the descent rate $-\dot V/V$ with
    both Jacobians and the band $[2k_t\beta,\,2k_t]$ (shaded) that
    \eqref{eq:Vdot} predicts for the damped inverse.}
    \label{fig:stability}
\end{figure}

We analyze a class of control laws that includes the damped inverse
\eqref{eq:damped_law}. In this class, the \emph{spectral filter} is a scalar
function $f:[0,\infty)\to\mathbb R$. The matrix $AA^{\mathsf T}$ is
symmetric and positive semidefinite and thus has the eigendecomposition
$AA^{\mathsf T}=U\operatorname{diag}(\lambda_1,\ldots,\lambda_{n_t})U^{\mathsf
T}$, where $n_t$ is the number of tip coordinates, $U$ is orthogonal, and the
eigenvalues $\lambda_i\ge0$ are the squared singular values of $A$. Applying
$f$ to each eigenvalue gives the matrix
\begin{equation}
    f(AA^{\mathsf T})
    =
        U\operatorname{diag}\bigl(f(\lambda_1),\ldots,f(\lambda_{n_t})\bigr)U^{\mathsf T},
    \label{eq:filter}
\end{equation}
which has the eigenvectors of $AA^{\mathsf T}$ and the eigenvalues
$f(\lambda_i)$. Each control law of the class computes $\bm\nu$ with this
matrix,
\begin{equation}
    \bm\nu
    =
    A^{\mathsf T}
    f(AA^{\mathsf T})
    Wv_{\mathrm{res}}.
    \label{eq:spectral_law}
\end{equation}
The damped inverse \eqref{eq:damped_law} corresponds to
$f(\lambda)=(\lambda+\xi^2)^{-1}$, for which $f(AA^{\mathsf
T})=(AA^{\mathsf T}+\xi^2I)^{-1}$, and $f(\lambda)=1$ gives the
Jacobian-transpose control law $\bm\nu=A^{\mathsf T}Wv_{\mathrm{res}}$.

\begin{assumption}[Nominal on-manifold operation]
\label{ass:nominal}
(i)~The target $p_D^\star$ is fixed; (ii)~the controller uses the exact tip
Jacobian $J_t$ of the plant; (iii)~the tendon active set is fixed, so that the
maps involved are smooth; (iv)~no barrier is active; and (v)~the EE
constraint holds, $\bm e_c=0$.
\end{assumption}

Under Assumption~\ref{ass:nominal}, $\dot{\bm q}_c=0$, and the command
\eqref{eq:hierarchical_law}, \eqref{eq:spectral_law} becomes
\begin{equation}
    \dot{\bm q}
    =
    N_cA^{\mathsf T}
    f(AA^{\mathsf T})
    W(k_t\bm e).
    \label{eq:on_manifold_controller}
\end{equation}
Since $N_c$ is symmetric and idempotent, $J_tN_cA^{\mathsf T}=AA^{\mathsf T}$,
and the tip error evolves as
\begin{equation}
    \dot{\bm e}
    =
    -k_t
    AA^{\mathsf T}
    f(AA^{\mathsf T})
    W\bm e.
    \label{eq:error_dynamics}
\end{equation}

\begin{proposition}[Output-error descent]
\label{prop:descent}
Let Assumption~\ref{ass:nominal} hold, and consider the weighted output-error
function, a Lyapunov candidate \cite{khalil2002nonlinear},
\begin{equation}
    V(\bm e)
    =
    \frac{1}{2}\bm e^{\mathsf T}W\bm e,
    \qquad
    W\succ0.
    \label{eq:lyapunov}
\end{equation}
If the spectral filter \eqref{eq:filter} is positive, that is, $f(\lambda_i)>0$
for every eigenvalue $\lambda_i$ of $AA^{\mathsf T}$, then
\begin{equation}
    \dot V
    =
    -k_t
    (W\bm e)^{\mathsf T}
    AA^{\mathsf T}
    f(AA^{\mathsf T})
    (W\bm e)
    \le 0,
    \label{eq:Vdot}
\end{equation}
and
\begin{equation}
    \dot V=0
    \quad\Longleftrightarrow\quad
    A^{\mathsf T}W\bm e=0.
    \label{eq:stationary_condition}
\end{equation}
\end{proposition}

\begin{IEEEproof}
By \eqref{eq:filter},
\[
    AA^{\mathsf T}f(AA^{\mathsf T})
    =
    U\operatorname{diag}\bigl(\lambda_1f(\lambda_1),\ldots,\lambda_{n_t}f(\lambda_{n_t})\bigr)U^{\mathsf T},
\]
which is symmetric and positive semidefinite since $\lambda_i\ge0$ and
$f(\lambda_i)>0$. Substituting \eqref{eq:error_dynamics} into $\dot V=\bm
e^{\mathsf T}W\dot{\bm e}$ gives \eqref{eq:Vdot}. The quadratic form in
\eqref{eq:Vdot} vanishes only on the null space of this matrix, which coincides
with the null space of $AA^{\mathsf T}$ and therefore with that of
$A^{\mathsf T}$, as stated in \eqref{eq:stationary_condition}.
\end{IEEEproof}

\begin{corollary}[Filter-independent stationary set]
\label{cor:stationary}
Under Assumption~\ref{ass:nominal}, every positive spectral filter has the same
stationary set,
\begin{equation}
    \mathcal E
    =
    \left\{
        \bm q:
        A(\bm q)^{\mathsf T}W\bm e(\bm q)=0
    \right\}.
    \label{eq:stationary_set}
\end{equation}
The filter therefore changes the trajectories and the convergence rates, but not
the set of stationary points.
\end{corollary}

At a stationary point, $W\bm e$ is orthogonal to $\operatorname{range}A$, that
is, the gradient of $V$ along the constraint manifold vanishes. Such a
point is not necessarily a local minimum of $V$, and its tip error can be
nonzero.
Note that multiple joint configurations can produce the same tip position, and
convergence of the tip error thus does not imply convergence of the joint
angles.

\begin{corollary}[Exponential convergence]
\label{cor:exponential}
Let Assumption~\ref{ass:nominal} hold in a region of the constraint manifold in
which $A$ has full row rank, that is, $\operatorname{rank}A=n_t$, and the
eigenvalues $\lambda_if(\lambda_i)$ of $AA^{\mathsf T}f(AA^{\mathsf T})$
satisfy $\min_i\lambda_if(\lambda_i)\ge\beta>0$. Then
\begin{equation}
    \dot V
    \le
    -2k_t\beta\lambda_{\min}(W)V
    \label{eq:exp_convergence}
\end{equation}
in this region. The tip error thus converges exponentially while the trajectory
remains in it.
\end{corollary}

\begin{IEEEproof}
By \eqref{eq:Vdot}, $\dot V\le-k_t\beta\|W\bm e\|^2$, and $\|W\bm e\|^2=\bm
e^{\mathsf T}W^2\bm e\ge\lambda_{\min}(W)\,\bm e^{\mathsf T}W\bm e
=2\lambda_{\min}(W)V$.
\end{IEEEproof}

For the damped inverse with $W=I$, under the conditions of
Corollary~\ref{cor:exponential}, the eigenvalues $\lambda_i/(\lambda_i+\xi^2)$ of
$AA^{\mathsf T}f(AA^{\mathsf T})$ lie in $[\beta,1)$, and \eqref{eq:Vdot} bounds
the descent rate as $2k_t\beta\le-\dot V/V<2k_t$. Fig.~\ref{fig:stability}
illustrates Proposition~\ref{prop:descent} and Corollary~\ref{cor:exponential}
on the simulated plant. With the exact Jacobian, $V$ decreases monotonically after each
target switch, faster than the bound \eqref{eq:exp_convergence}, and its descent
rate remains in this band up to small deviations caused by the discrete time
steps of the simulation. The identified
model \eqref{eq:JD_identified} used by the implemented controller violates
condition~(ii) of Assumption~\ref{ass:nominal}, and the descent rate then leaves
the band by up to $8\,\%$, but $V$ still decreases monotonically.

\begin{figure*}[t]
    \centering
    \includegraphics[width=\textwidth]{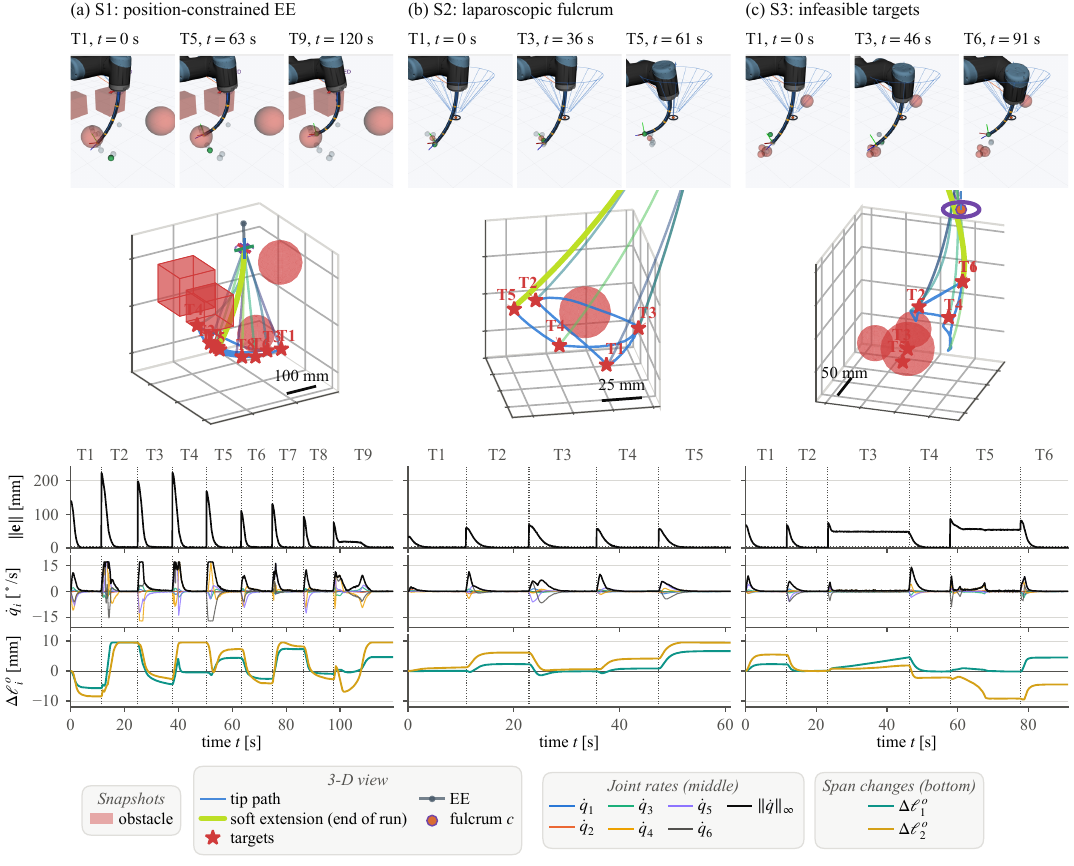}
    \caption{The three simulated scenarios: (a) position-constrained EE, (b)
    laparoscopic fulcrum, and (c) infeasible targets. Top: snapshots at the
    start and end of two target intervals. Middle: 3-D view with the targets,
    the obstacles, the tip path, and the soft extension at the end of each
    target (thin) and of the run (thick). Bottom: norm of the tip error with the
    $4\,\mathrm{mm}$ arrival tolerance, joint rates, and changes of the two
    exterior spans; dotted lines mark target changes. In (c), T3 lies inside an
    obstacle and T5 beyond the insertion limit, and the error settles at a
    nonzero value while the corresponding barrier is active.}
    \label{fig:sim_all}
\end{figure*}

\section{Simulation Results}
\label{sec:results}
\subsection{Setup}
\label{subsec:sim_setup}

\begin{table}[t]
    \centering
    \caption{Simulation setup.}
    \label{tab:sim_setup}
    \footnotesize
    \setlength{\tabcolsep}{3pt}
    \renewcommand{\arraystretch}{1.12}
    \begin{tabularx}{\columnwidth}{@{}l>{\raggedright\arraybackslash}X@{}}
        \toprule
        Plant & UR10 with a $320\,\mathrm{mm}$ Cosserat rod
            \cite{tummers2023cosserat} and two tendons, $100\,\mathrm{Hz}$ \\
        Sensing & Joint states; tip pose and five body points at
            $20\,\mathrm{Hz}$, Gaussian noise of $0.5\,\mathrm{mm}$ standard
            deviation \\
        Identification & $J_{\ell q}$, $J_{\mathrm{transport}}$ exact; $\widehat
            J_{\mathrm{span}}$ fitted on an $11\times11$ grid of spans
            ($\pm10\,\mathrm{mm}$), refined online \\
        Control & $50\,\mathrm{Hz}$; $k_t=1.4$, $k_c=3.0$,
            $\xi=\mu=0.04\,\mathrm{m}$; joint rates
            $\le17^\circ/\mathrm{s}$ \\
        Targets & Reached when the tip stays within $4\,\mathrm{mm}$ for
            $8\,\mathrm{s}$ \\
        \bottomrule
    \end{tabularx}
\end{table}

At each time step of the simulation, the plant solves the quasi-static
equilibrium of a Cosserat rod \cite{tummers2023cosserat} loaded by the two
tendons, while the controller uses the identified model of
Section~\ref{subsec:identified}. The controller receives only the joint states
and noisy 3-D measurements of the tip and five body points. We log the
noise-free simulator state separately and use it only for evaluation.
Table~\ref{tab:sim_setup} summarizes the simulation setup.

Within the identified span domain ($\pm10\,\mathrm{mm}$), the RMS tip-position
error on validation points not used in the fit is $1.5\,\mathrm{mm}$.
Extending the evaluation range to $\pm22\,\mathrm{mm}$ increases this error to
$5.5\,\mathrm{mm}$. We therefore restrict the spans to the identified span
domain with the span-domain barrier $h_{\mathrm{span}}$. For the obstacle
barrier, the clearance is computed to the polyline through the five body points.
These points are $77$--$90\,\mathrm{mm}$ apart, and the soft extension can
deviate from this polyline between them. The barrier thus subtracts a constant
margin that exceeds the largest such deviation measured within the identified
span domain. In simulation, the controller solves the QP \eqref{eq:safety_qp} at
every control step.

\subsection{Scenarios}
\label{subsec:sim_scenarios}

\begin{table}[t]
    \centering
    \caption{Simulated scenarios. All impose the obstacle and span-domain
    barriers, and the fulcrum scenarios also impose the cone and insertion
    barriers.}
    \label{tab:sim_scenarios}
    \footnotesize
    \setlength{\tabcolsep}{3.5pt}
    \renewcommand{\arraystretch}{1.12}
    \begin{tabular}{@{}llccl@{}}
        \toprule
        & EE constraint & Obst. & Targets & Infeasible by design \\
        \midrule
        S1 & position-constrained & 4 & 9 & -- \\
        S2 & laparoscopic fulcrum & 1 & 5 & -- \\
        S3 & laparoscopic fulcrum & 4 & 6 & T3, T5 \\
        \bottomrule
    \end{tabular}
\end{table}

In scenario S1 (Table~\ref{tab:sim_scenarios}, Fig.~\ref{fig:sim_all}a), the EE
position is fixed and its orientation is unconstrained (position-constrained EE,
Table~\ref{tab:ee_constraints}). The tip therefore moves only through the
reorientation of the EE and the deformation of the soft extension, both produced
by the three directions of null-space motion. Its nine targets lie around two
spheres and two boxes.

Scenario S2 (Fig.~\ref{fig:sim_all}b) uses the laparoscopic fulcrum
(Table~\ref{tab:ee_constraints}) in the configuration of
Fig.~\ref{fig:mechanism}. Five inspection points lie below the aperture, and a
small sphere representing a polyp blocks the direct path between two of them.

In scenario S3 (Fig.~\ref{fig:sim_all}c), we evaluate the response of the
controller to targets that are infeasible by design. This scenario reuses the
aperture and four targets of S2, places four spheres along the route, and adds
two infeasible targets, T3 inside one of the spheres and T5 beyond the insertion
limit ($h_{\mathrm{ins}}$ in Table~\ref{tab:barriers}). In all scenarios, the
run advances to the next target if the tip error does not decrease for
$10\,\mathrm{s}$.

\subsection{Target Regulation and Constraint Satisfaction}
\label{subsec:sim_results}

All targets classified as feasible meet a $4\,\mathrm{mm}$ arrival error criterion
in the three scenarios (Fig.~\ref{fig:sim_all}, top row of time plots). After
each change to a feasible target, the tip error decreases without sustained
oscillation. The joint rates (Fig.~\ref{fig:sim_all}, middle row) increase
mainly after target changes and remain within the rate limit. The geometric deviation from the EE
constraint
remains below $0.2\,\mathrm{mm}$ during the runs. The EE position stays within
$0.18\,\mathrm{mm}$ of its reference in S1 (Fig.~\ref{fig:sim_all}a), and the
tool axis stays within $0.04\,\mathrm{mm}$ of the fulcrum point in S2 while the
arm pivots and inserts the soft extension (Fig.~\ref{fig:sim_all}b).

In all scenarios, the barrier values remain nonnegative. To isolate the effect
of the obstacle barrier, we repeated a run with a sphere on the tip path while
the barrier was monitored but not imposed (not shown in Fig.~\ref{fig:sim_all}).
The soft extension then penetrated the sphere by $18.8\,\mathrm{mm}$, whereas
with the barrier imposed, the clearance remained above $10.7\,\mathrm{mm}$ and
the target was still reached. Around the targets of S2, however, the span domain
bounds the reachable tip positions more tightly than the obstacle, cone, and
insertion barriers.

\subsection{Infeasible Targets}
\label{subsec:sim_limits}

For the two infeasible targets of S3, the tip error settles at a nonzero value
once a barrier becomes active and prevents its further reduction. The obstacle
barrier becomes active with the tip $48\,\mathrm{mm}$ from T3, and the EE
reaches the insertion limit with the tip $55\,\mathrm{mm}$ from T5. After the
run advances, the remaining targets are reached.

\begin{figure}[t]
    \centering
    \includegraphics[width=\columnwidth]{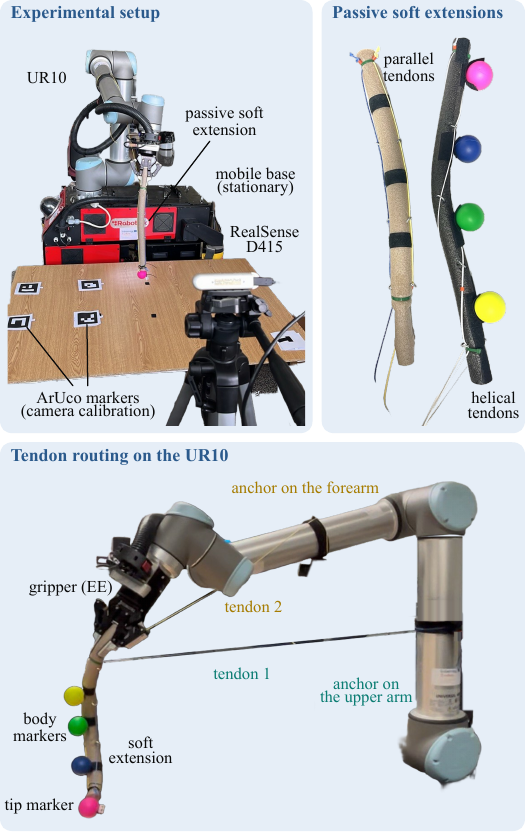}
    \caption{Experimental platform. Top left: the UR10 on a stationary mobile
    base, the passive soft extension, and the RGB-D camera, calibrated in the
    robot base frame with the ArUco markers on the table. Top right: passive
    soft extensions with parallel tendons (Exps.~1--5) and with helical tendons
    of one turn each (Exp.~6); colored balls mark three body points and the tip.
    Bottom: tendon routing from anchors on the upper arm (tendon~1) and the
    forearm (tendon~2) to the base of the soft extension at the gripper.}
    \label{fig:exp_setup}
\end{figure}

\begin{figure*}[t]
    \centering
    \includegraphics[width=\textwidth]{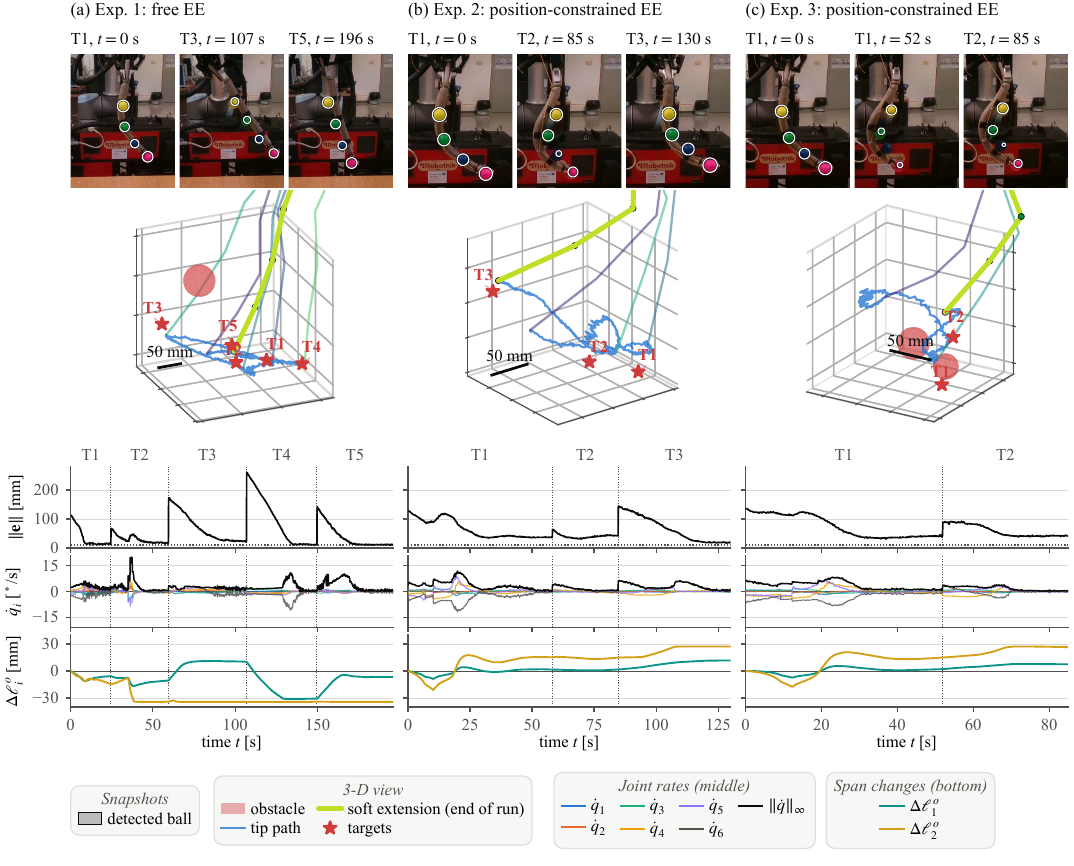}
    \caption{Experiments with (a) a free EE and (b, c) a position-constrained
    EE. Top: camera snapshots with the detected balls highlighted. Middle: 3-D
    view with the targets, the obstacles, the tip path, and the soft extension
    at the end of each target (thin) and of the run (thick). Bottom: norm of the
    tip error with the $10\,\mathrm{mm}$ threshold, joint rates, and changes of
    the two exterior spans; dotted lines mark target changes. The time plots
    have the same scales as in Fig.~\ref{fig:exp_fulcrum}.}
    \label{fig:exp_rigid}
\end{figure*}

\section{Experimental Results}
\label{sec:experiments}

\begin{figure}[t]
    \centering
    \includegraphics[width=\columnwidth]{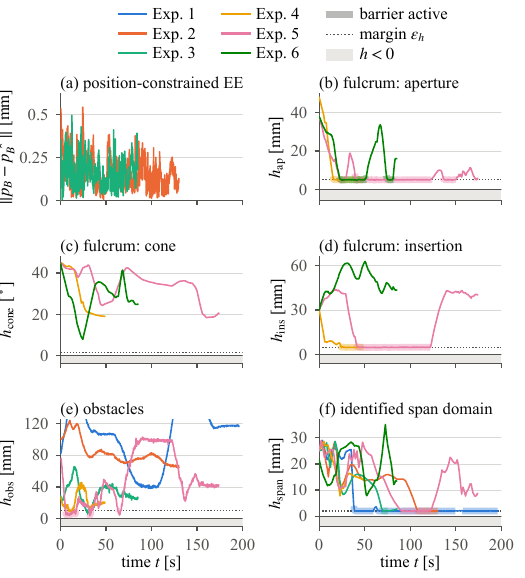}
    \caption{EE constraint and barrier values (Table~\ref{tab:barriers})
    measured in the six experiments. Shading marks negative barrier values,
    dotted lines the margins $\varepsilon_h$, and thick bands the intervals in
    which the barrier condition was active in the QP. (a)~Distance of the EE
    from its reference for the position-constrained EE. (b)--(d)~Aperture,
    cone, and insertion barriers of the fulcrum. (e)~Clearance of the tracked
    body to the obstacles, clipped at $125\,\mathrm{mm}$, with the
    $10\,\mathrm{mm}$ margin below which the obstacle condition is imposed.
    (f)~Smallest distance of the spans to the bounds of the identified span
    domain.}
    \label{fig:exp_constraints}
\end{figure}

\begin{figure*}[t]
    \centering
    \includegraphics[width=\textwidth]{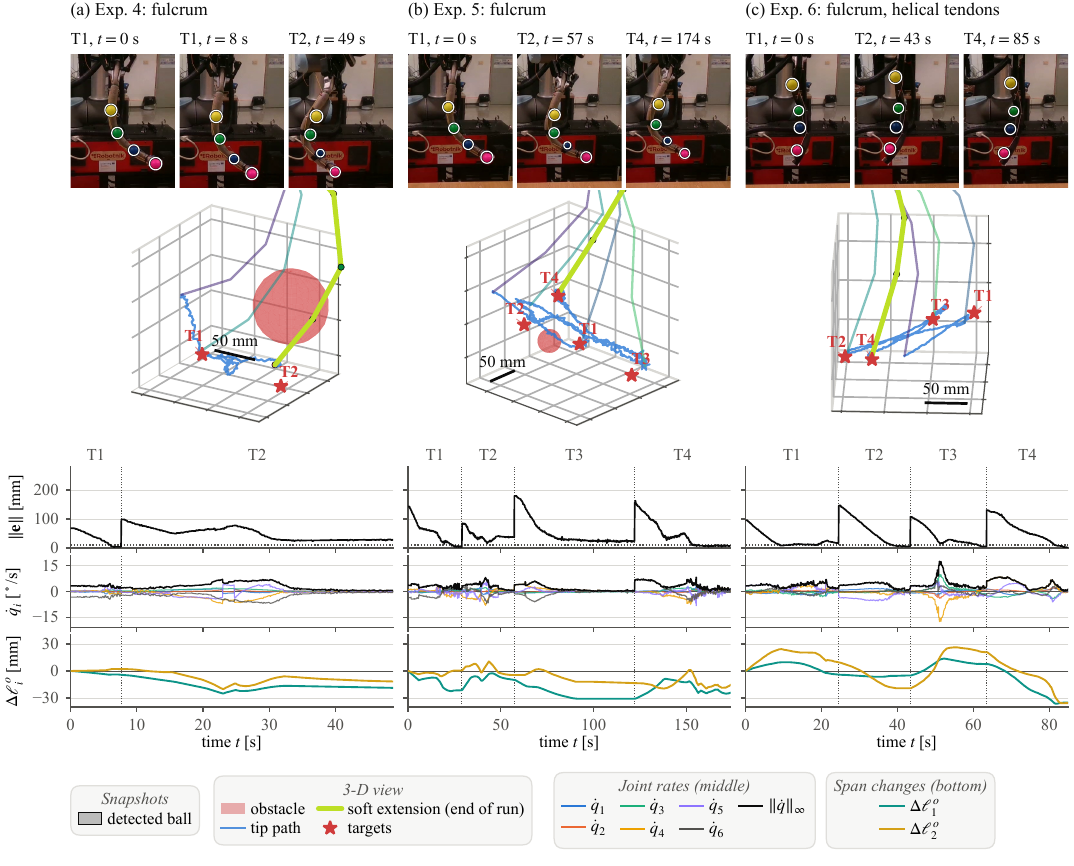}
    \caption{Experiments with the fulcrum at a finite aperture, with parallel
    tendons (a, b) and with helical tendons (c). Each column shows snapshots, a
    3-D view, the tip error, the joint rates, and the span changes in the layout
    of Fig.~\ref{fig:exp_rigid}. In (b), T3 is infeasible by design.}
    \label{fig:exp_fulcrum}
\end{figure*}

\subsection{Setup}
\label{subsec:exp_setup}

\begin{table}[t]
    \centering
    \caption{Experimental setup.}
    \label{tab:exp_setup}
    \footnotesize
    \setlength{\tabcolsep}{3pt}
    \renewcommand{\arraystretch}{1.12}
    \begin{tabularx}{\columnwidth}{@{}l>{\raggedright\arraybackslash}X@{}}
        \toprule
        Plant & UR10; passive foam tube, $475\,\mathrm{mm}$ long, less than
            US\$10 each; two tendons, anchored at mid upper arm and mid
            forearm \\
        Sensing & Joint states; RealSense D415 at $30\,\mathrm{Hz}$, three body
            markers and a tip marker \\
        Identification & Linear $\widehat J_{\mathrm{span}}$ from $13$ probe
            poses ($10$ for fitting, $3$ for validation), refined online \\
        Control & Joint velocity commands at $125\,\mathrm{Hz}$,
            $8$--$12^\circ/\mathrm{s}$ at most; $k_t=1.2$,
            $\xi=0.02\,\mathrm{m}$, $\mu=0.04\,\mathrm{m}$; tip speed
            $\le20\,\mathrm{mm/s}$ \\
        Targets & Reached when the steady-state tip error is below
            $10\,\mathrm{mm}$ \\
        \bottomrule
    \end{tabularx}
\end{table}

The two tendons are anchored on the upper arm and on the forearm
(Fig.~\ref{fig:exp_setup}). Both soft extensions are handmade from foam
tubing with cord tendons, which run through small eyelets along the tube and are
fastened with cable ties. A stationary RGB-D camera calibrated in the robot base frame measures the
positions of the colored balls that mark three body points and the tip
(Fig.~\ref{fig:exp_setup}). From these measurements at the probing poses, we identify a
linear $\widehat J_{\mathrm{span}}$ for the model
\eqref{eq:JD_identified} (Table~\ref{tab:exp_setup}), with separate models for
Exp.~1, for Exps.~2--5, and for the helical soft extension of Exp.~6. Across the three validation
poses, their RMS
errors over the coordinates of all four markers are $9$--$14\,\mathrm{mm}$.
These errors are not directly comparable to the $1.5\,\mathrm{mm}$ simulation
error, which is computed for the tip position alone. The controlled output is
the 12-dimensional vector of the positions of the three
body markers and the tip, with $W$ assigning the largest weight to the tip.

\subsection{Experimental Protocol}
\label{subsec:exp_runs}

\begin{table}[t]
    \centering
    \caption{Experiments. All impose the span-domain barrier and, when
    obstacles are present, the obstacle barrier. Under the fulcrum, they also
    impose the three fulcrum barriers (aperture, cone, and insertion). Targets:
    number predicted feasible
    ${+}$ number infeasible by design.}
    \label{tab:exp_runs}
    \footnotesize
    \setlength{\tabcolsep}{4pt}
    \begin{tabular}{@{}cllcc@{}}
        \toprule
        Exp. & EE constraint & Tendons & Obst. & Targets \\
        \midrule
        1 & free & parallel & 1 & $3+2$ \\
        2 & position-constrained & parallel & 1 & $2+1$ \\
        3 & position-constrained & parallel & 2 & $2$ \\
        4 & laparoscopic fulcrum & parallel & 1 & $2$ \\
        5 & laparoscopic fulcrum & parallel & 1 & $3+1$ \\
        6 & laparoscopic fulcrum & helical & 0 & $4$ \\
        \bottomrule
    \end{tabular}
\end{table}

Table~\ref{tab:exp_runs} lists the six experiments, with a free EE, a
position-constrained EE, and a fulcrum at a finite aperture. This aperture is
a circle of radius $r_{\mathrm{ap}}$, and the tool axis may cross the aperture
plane anywhere inside it. Unlike the ideal fulcrum
of Table~\ref{tab:ee_constraints} ($r_c=2$), this fulcrum therefore has no
equality part ($r_c=0$) and is imposed only through the aperture, cone, and
insertion barriers of Table~\ref{tab:barriers}, with a cone half-angle of
$45^\circ$.

Each experiment commands a sequence of two to five tip targets near the initial
tip position. Most targets are generated from the identification data and the
identified model, with their predicted spans inside $\mathcal L$. A few are
infeasible by design to evaluate the controller when barriers prevent
convergence. During a run, the target sequence advances when the tip error stays
below a $6\,\mathrm{mm}$ arrival tolerance for $1.5\,\mathrm{s}$ or does not
improve by $0.5\,\mathrm{mm}$ within $15\,\mathrm{s}$. Independently of how the
run advances, a target counts as reached in the results when its steady-state
tip error is below $10\,\mathrm{mm}$, about $2\,\%$ of the length of the soft
extension.

Unlike in simulation, the nominal command \eqref{eq:hierarchical_law} is sent
unchanged while it satisfies every CBF condition of \eqref{eq:safety_qp}, and
the QP is solved otherwise. The obstacles are virtual spheres, and their
clearance is the distance to the polyline through the EE and the tracked markers
minus the radius of the soft extension. Once the clearance falls below its
$10\,\mathrm{mm}$ margin, the obstacle condition is added in the form
$\nabla h_{\mathrm{obs}}^{\mathsf T}\dot{\bm q}\ge0$, which requires the
clearance rate predicted by the model to be nonnegative.

\subsection{Actuation and Constraint Satisfaction}
\label{subsec:exp_results}

The experiments with a free EE and a position-constrained EE
(Fig.~\ref{fig:exp_rigid}) show how the contributions of transport and
deformation to the tip motion depend on the EE constraint. With the EE free
(Exp.~1, Fig.~\ref{fig:exp_rigid}a), most of the tip motion is transport. The
EE moves up to $160\,\mathrm{mm}$ from its initial position. Both spans
shorten as the arm releases the tendons, and the soft extension straightens
slightly. With a position-constrained EE (Exps.~2 and~3,
Fig.~\ref{fig:exp_rigid}b--c), the EE stays within $0.5\,\mathrm{mm}$ of its
reference (Fig.~\ref{fig:exp_constraints}a), and the tip moves only through EE
reorientation and deformation of the soft extension. Span~2 lengthens by up to $27\,\mathrm{mm}$ in
these experiments as the arm
pulls on tendon~2, which
produces the largest deformation of all six experiments (snapshots in
Fig.~\ref{fig:exp_rigid}b--c). Across all experiments, the joint rates are generally below
$10^\circ/\mathrm{s}$, and the spans change by up to about $35\,\mathrm{mm}$
(Figs.~\ref{fig:exp_rigid} and~\ref{fig:exp_fulcrum}, bottom).

The measured barrier values remain nonnegative in all six experiments
(Fig.~\ref{fig:exp_constraints}). Under the fulcrum (Exps.~4--6,
Fig.~\ref{fig:exp_constraints}b--d), the measured tool axis remains inside the
aperture, and the aperture barrier is active for more than half of the duration
of Exps.~4 and~5. The insertion depth also reaches its limit, where the
insertion barrier becomes active, whereas the tilt of the tool axis remains
below $38^\circ$, within the $45^\circ$ half-angle of the cone. Unlike the fulcrum conditions, the
obstacle condition
(Fig.~\ref{fig:exp_constraints}e) relies on the identified body-point Jacobians,
linearized over each control step. These approximations
allow the clearance to decrease below its $10\,\mathrm{mm}$ activation margin
without crossing the zero boundary of the obstacle barrier, reaching a minimum of
$3.9\,\mathrm{mm}$ in Exp.~5. The soft extension bends between the two obstacles of Exp.~3 and stays
at
least $6\,\mathrm{mm}$ from both. Finally, the span-domain barrier (Fig.~\ref{fig:exp_constraints}f)
becomes active in
Exps.~1, 2, 3, and~5, with the spans at the boundary of the identified span
domain.

After a new calibration and identification, we evaluate the same controller with
helical routing in Exp.~6 (Fig.~\ref{fig:exp_fulcrum}c). The controller reaches
three of the four targets. Both exterior spans lengthen and shorten
over a wider range than in the experiments with parallel tendons.
These results suggest that tendon routing influences the span variations
available from arm motion.

\subsection{Reached and Unreached Targets}
\label{subsec:exp_missed}

\begin{table}[t]
    \centering
    \caption{Unreached targets, grouped by limiting factor, with the
    steady-state tip error. The barriers are those of
    Table~\ref{tab:barriers}. Stationary points are possible since the output
    has 12 coordinates, whereas the rank of $A$ is at most six
    (Section~\ref{subsec:exp_missed}).}
    \label{tab:exp_missed}
    \footnotesize
    \setlength{\tabcolsep}{3pt}
    \renewcommand{\arraystretch}{1.15}
    \begin{tabularx}{\columnwidth}{@{}>{\raggedright\arraybackslash}p{2.75cm}>{\raggedright\arraybackslash}X@{}}
        \toprule
        Limiting factor & Exp.--target (error in mm) \\
        \midrule
        Infeasible by design & 1--T2 (18), 1--T4 (12), 2--T2 (43), 5--T3 (24)
        \\
        Span domain & 1--T3 (23), 1--T5 (12), 2--T3 (19) \\
        Aperture and insertion & 4--T2 (28), 5--T2 (37), 6--T1 (17) \\
        Stationary point (no active barrier) & 1--T1 (14), 2--T1 (37), 3--T1
        (42), 3--T2 (42) \\
        \bottomrule
    \end{tabularx}
\end{table}

The tip error decreases and settles without sustained oscillation after each
target change in every experiment (Figs.~\ref{fig:exp_rigid}
and~\ref{fig:exp_fulcrum}, top row of time plots). Six of the 16 targets predicted feasible
are reached, all under the fulcrum, and Table~\ref{tab:exp_missed} groups the
remaining ones by their limiting factor.

For the targets that are infeasible by design, the tip approaches each target
until one or more active barriers prevent further reduction of the tip error.
In Exp.~5
(Fig.~\ref{fig:exp_fulcrum}b), for example, the tip settles $24\,\mathrm{mm}$
from T3, with the aperture, insertion, and span-domain barriers active, and then
reaches T4. Regarding the targets limited by the span domain, the spans converge to the margin of
the span-domain barrier (Fig.~\ref{fig:exp_constraints}f) while a nonzero tip
error remains. These targets were generated with the identified model, whose
prediction error is of the order of $1\,\mathrm{cm}$, and a target with predicted
spans inside $\mathcal L$ can thus require actual spans slightly outside $\mathcal
L$. For the targets limited by the aperture and insertion barriers, the tool axis
reaches the rim of the aperture and the EE reaches the end of its insertion
range (Fig.~\ref{fig:exp_constraints}b and~d).

Four targets remain unreached without an active barrier
(Fig.~\ref{fig:exp_rigid}a--c), which is consistent with the stationary points
with nonzero tip error permitted by \eqref{eq:stationary_condition}. Such points can
occur here as the output has 12 coordinates and the arm six joints, so that
$\operatorname{rank}A\le6<n_t$ and Corollary~\ref{cor:exponential} does not
apply. Proposition~\ref{prop:descent}, however, does not require
$\operatorname{rank}A=n_t$ and,
under Assumption~\ref{ass:nominal}, still guarantees that $V$ does not increase
for the 12-dimensional marker-position output.

\section{Conclusion}
\label{sec:conclusion}

We presented a passive tendon-driven soft extension actuated and regulated by
the joint motion of a rigid manipulator. Since its tendons are anchored upstream of the
EE, null-space motion can actuate them when an EE constraint restricts the
joint motion. Our differential model combines analytic arm and tendon kinematics
with a locally identified span-to-deformation map. The controller regulates the
tip while
control barrier functions impose obstacle-clearance, fulcrum, and model-validity
constraints.

The rank analysis gives a local bound on the number of independent deformation
directions that null-space motion can transmit through the tendons. Under
Assumption~\ref{ass:nominal}, the nominal control law does not increase the
weighted tip error, and the tip error converges exponentially in regions where
the rank and spectral conditions of Corollary~\ref{cor:exponential} hold. In
particular, these conditions require no more output coordinates than
independent null-space joint velocities, since the rank of $A$ cannot exceed
their number.

In simulation, the controller reached every target classified as feasible under
the position and fulcrum constraints. On the real setup, joint motion changed the exterior tendon
spans by up to
$35\,\mathrm{mm}$ in six experiments with a UR10 and handmade soft extensions
costing less than US\$10 each. Unreached targets were
associated with active barriers or with stationary points of the controller,
where the command vanishes although the tip error is nonzero. All barrier values
remained nonnegative, both in simulation and on the real setup. On the
validation poses, the identified models predicted the four marker positions with
RMS errors of $9$--$14\,\mathrm{mm}$. The experiments show that the tip and
body points of passive soft extensions can be regulated in closed loop through
the joint motion of a rigid manipulator, without dedicated tendon actuators.

\subsection{Limitations and Future Work}
\label{subsec:limitations}

The span-to-deformation map identified from probing motions is local to the
identified span domain $\mathcal L$. A mechanics-based model calibrated from the same data
could support prediction and planning beyond this domain, provided that its
extrapolation accuracy is validated.

We did not optimize the number of tendons or the placement of their anchors in
this paper. Both could be chosen for a given manipulator, tendon routing, and EE
constraint by maximizing a transmission measure derived from $J_{\ell q}N_c$.
The wider span
variations obtained with helical tendons in Exp.~6 also motivate a controlled
comparison of tendon routings.

Our controller regulates the geometry of the soft extension but not its
tendon tensions, contact forces, or stiffness. Identifying the span-to-tension
map $C_\tau$, together with force sensing or calibrated contact models, could
extend the controller to force or stiffness objectives when sufficient joint redundancy
remains after the geometric task.

\raggedbottom
\appendix[Notation]

The symbols used in the paper are listed below by topic, with the equation that
defines each of them where applicable.

\newcommand{\notationgroup}[2]{\par\noindent
  {\footnotesize\setlength{\tabcolsep}{3pt}\renewcommand{\arraystretch}{1.12}\begin{tabularx}{\columnwidth}{@{}>{$}l<{$}>{\raggedright\arraybackslash}X@{}}
  \toprule
  \multicolumn{2}{@{}l}{\emph{#1}}\\
  \midrule
  #2
  \bottomrule
  \end{tabularx}}\par\vspace{6pt}}

\notationgroup{Operators}{\|x\| & Euclidean norm of a vector $x$ \\
  {[x]_\times} & skew-symmetric matrix with $[x]_\times y=x\times y$ \\
  J^+ & Moore--Penrose pseudoinverse of $J$ \\
  \operatorname{rank},\ \operatorname{range} & rank and column space of a matrix \\
  \lambda_{\min} & smallest eigenvalue of a symmetric matrix \\
  M\succ0 & $M$ is positive definite \\
  \operatorname{diag},\ I & (block-)diagonal matrix; identity matrix \\
  \operatorname{dist}(p,\mathcal O) & distance from a point $p$ to a set $\mathcal O$ \\
}

\notationgroup{Arm and tendons}{n,\ \bm q & number of joints; joint angles \\
  \{B\},\ T_B=(R_B,p_B) & EE frame; its pose (rotation, position) \\
  J_B=(J_B^v,J_B^\omega) & geometric Jacobian of the EE (rows: velocity of
    $p_B$, angular velocity of $\{B\}$; world frame) \\
  m,\ i & number of tendons; tendon index \\
  a_i,\ J_{a_i}^v & anchor of tendon $i$ on the arm; its translational Jacobian \\
  {}^{B}b_i,\ b_i & entry point of tendon $i$ in $\{B\}$; in the world
    \eqref{eq:span} \\
  d_i & unit vector from $b_i$ to $a_i$ \\
  \ell^o_i,\ \bm\ell^o & exterior span of tendon $i$ \eqref{eq:span}; vector of
    spans \\
  J_{\ell q} & Jacobian from joint rates to span rates \eqref{eq:Jlq} \\
}

\notationgroup{Soft extension and model}{\{D\},\ p_D & distal frame; tip position \\
  {}^{B}\mathcal V_{D/B} & twist of $\{D\}$ relative to $\{B\}$, expressed in $\{B\}$ \\
  \bm\tau,\ \bm z & tendon tensions; shape coordinates (generalized coordinates
    of the deformed soft extension) \\
  K_a,\ m_a & diagonal stiffness matrix of the tendons, zero for slack or
    saturated tendons; number of active tendons \\
  J_{\ell z},\ J_{z\tau} & Jacobian from shape coordinates to interior tendon
    lengths; Jacobian from tensions to shape coordinates \\
  C_\tau,\ G & span-to-tension map \eqref{eq:Ctau}; compliance of the soft
    extension seen at the tendons \\
  J_{D\tau} & Jacobian from tensions to the relative distal twist \\
  J_{\mathrm{span}},\ \widehat J_{\mathrm{span}} & span-to-deformation map
    \eqref{eq:Jspan}; its identified estimate \\
  J_{\mathrm{transport}},\ \bar R_B & Jacobian of $\{D\}$ at fixed shape
    \eqref{eq:Jtransport}; $\operatorname{diag}(R_B,R_B)$ \\
  J_D,\ \widehat J_D & distal-twist Jacobian \eqref{eq:JD}; its identified
    model \eqref{eq:JD_identified} \\
  S,\ J_t & selector of the translational rows; tip-position Jacobian
    $J_t=SJ_D$ \\
  p_j,\ {}^{B}p_j,\ J_{p_j} & body points in the world and in $\{B\}$; their
    Jacobian \eqref{eq:shape_chain} \\
  \mathcal L=[\bm\ell^-,\bm\ell^+] & identified span domain \\
  J_{\mathrm{def}} & constrained deformation Jacobian \eqref{eq:Jdef} \\
}

\notationgroup{Controller}{\bm e_c,\ J_c,\ r_c & EE-constraint error
(Table~\ref{tab:ee_constraints}),
    Jacobian \eqref{eq:constraint_error_rate}, and rank \\
  p_B^\star,\ c,\ u & held EE point; fulcrum point; tool axis
    (Table~\ref{tab:ee_constraints}) \\
  N_c & projector onto the null space of $J_c$ \eqref{eq:Nc} \\
  \dot{\bm q}_c,\ k_c & constraint-restoring velocity \eqref{eq:primary}; its gain \\
  \dot{\bm q}_{\mathrm{nom}} & nominal command \eqref{eq:hierarchical_law} \\
  p_D^\star,\ \bm e & tip target; tip error \eqref{eq:tip_error} \\
  v^\star,\ k_t & desired tip velocity; tip gain \eqref{eq:tip_error} \\
  A & constrained tip Jacobian $J_tN_c$ \eqref{eq:A} \\
  \bm\nu,\ v_{\mathrm{res}} & joint-space vector in the null-space command
    \eqref{eq:damped_law}; residual tip velocity \\
  \xi,\ W & damping of the DLS inverse \eqref{eq:damped_law}; weight of the tip
    coordinates \\
}

\notationgroup{Safety program and analysis}{h\in\mathcal H,\ \varepsilon_h,\ \alpha_h & barrier
function in the set of
    barriers (Table~\ref{tab:barriers}); its margin and CBF rate
    \eqref{eq:safety_qp} \\
  w_c,\ \mu,\ \dot{\bm q}^{\pm} & weight of the EE task in \eqref{eq:safety_qp};
    regularization weight of the joint rates; joint-rate limits \\
  \mathcal O,\ r,\ r_{\mathrm{ap}} & obstacle; radii of the soft extension and
    of the aperture \\
  \mathcal P & polyline through the body points \\
  \eta,\ J_p & unit vector from $\mathcal O$ to its closest point $p$ on
    $\mathcal P$; Jacobian of $p$ \\
  d_c,\ \theta,\ \bar\theta & distance from $c$ to the tool axis; tilt of the
    tool axis; half-angle of the cone \\
  \delta,\ \delta_{\min},\ \delta_{\max} & insertion depth and its limits \\
  f,\ f(AA^{\mathsf T}) & spectral filter (a scalar function); its matrix form
    \eqref{eq:filter} \\
  U & orthogonal matrix of eigenvectors of $AA^{\mathsf T}$ \\
  \lambda_i,\ n_t & eigenvalues of $AA^{\mathsf T}$ (squared singular values of
    $A$); number of tip coordinates \\
  V,\ \beta & weighted output-error (Lyapunov) function \eqref{eq:lyapunov};
    lower bound on the eigenvalues $\lambda_if(\lambda_i)$ of $AA^{\mathsf
    T}f(AA^{\mathsf T})$ in Corollary~\ref{cor:exponential} \\
  \mathcal E & stationary set \eqref{eq:stationary_set} \\
}

\section*{Acknowledgment}

Anthropic's Claude was used to assist with the implementation and debugging of
the simulation and experiment software
(Sections~\ref{sec:analysis}--\ref{sec:experiments}) and of the code that
generates the figures and the supplementary video. Claude was also used to draft
parts of the manuscript text, and Claude and OpenAI's ChatGPT to revise it,
under the authors' direction. The
authors directed the research, checked all AI-assisted code, text, and figures,
and retain full responsibility for the methods, results, and manuscript content.

\bibliographystyle{IEEEtran}
\bibliography{references}

\end{document}